%% file: main_paper.tex
\documentclass[10pt,journal,compsoc]{IEEEtran}

\ifCLASSOPTIONcompsoc
  \usepackage[nocompress]{cite}
\else
  \usepackage{cite}
\fi

\ifCLASSINFOpdf
  \usepackage[pdftex]{graphicx}
  \graphicspath{{../figures/}}
  \DeclareGraphicsExtensions{.pdf}
\else
  \usepackage[dvips]{graphicx}
  \graphicspath{{../figures/}}
  \DeclareGraphicsExtensions{.pdf}
\fi

\usepackage{amssymb} 
\usepackage{amsmath}
\usepackage{array}
\usepackage{subfig}
\usepackage{booktabs}
\usepackage{algorithm}  
\usepackage{algorithmicx}
\usepackage{algpseudocode}  
\usepackage{multirow}
\usepackage{enumerate}
\usepackage{xcolor}
\usepackage{comment}
\usepackage{colortbl}
\usepackage{hyperref}
\usepackage{xcolor}    
\usepackage{pifont} 

\usepackage{amsmath} 
\usepackage{caption} 

\usepackage[capitalize]{cleveref}
\crefname{section}{Sec.}{Secs.}
\Crefname{section}{Section}{Sections}
\Crefname{table}{Table}{Tables}
\crefname{table}{Tab.}{Tabs.}

\newcommand{\revise}[1]{\textcolor{black}{#1}}
\newcommand{\re}[1]{\textcolor{black}{#1}}
\newcommand{\rv}[1]{\textcolor{black}{#1}}

\begin{document}
\title{TagCLIP: Improving Discrimination Ability of Zero-Shot Semantic Segmentation}

\author{Jingyao~Li,
        Pengguang~Chen,
        Shengju~Qian,
        Shu~Liu,~\IEEEmembership{Member,~IEEE}
        and~Jiaya~Jia,~\IEEEmembership{Fellow,~IEEE}
\IEEEcompsocitemizethanks{\IEEEcompsocthanksitem Jingyao Li and Shengju Qian are with the Department of Computer Science and Engineering of the Chinese University of Hong Kong (CUHK) \\
Jiaya Jia's E-mail: leojia9@gmail.com
\IEEEcompsocthanksitem Pengguang Chen, Shu Liu and Jiaya Jia are with SmartMore.}
\thanks{Manuscript received Sep 3rd, 2024.}}


\IEEEtitleabstractindextext{%
\input{sections/01_abstract.tex}
}

\maketitle
\IEEEdisplaynontitleabstractindextext
\IEEEpeerreviewmaketitle

\IEEEraisesectionheading{\section{Introduction}\label{sec:intro}}
\input{sections/02_intro.tex}

\section{Related Works}\label{sec:related}
\input{sections/03_relatex_works.tex}

\section{Methods}\label{sec:methods}
\input{sections/04_methods.tex}

\section{Experiments}\label{sec:exp}
\input{sections/05_experiment.tex}

\section{Conclusion}\label{sec:conclusion}
\input{sections/08_conclusion.tex}

\ifCLASSOPTIONcompsoc
\else
\fi


\ifCLASSOPTIONcaptionsoff
  \newpage
\fi

\bibliographystyle{IEEEtran}
\bibliography{egbib}

\appendices
\input{sections/09_bio.tex}

\end{document}

%% file: sections/01_abstract.tex
\begin{abstract}
 \revise{Contrastive Language-Image Pre-training (CLIP) has recently shown great promise in pixel-level zero-shot learning tasks.} However, existing approaches utilizing CLIP's text and patch embeddings to generate semantic masks often misidentify input pixels from unseen classes, leading to confusion between novel classes and semantically similar ones. In this work, we propose a novel approach, \textbf{TagCLIP}~(\textbf{T}rusty-\textbf{a}ware \textbf{g}uided CLIP), to address this issue. We disentangle the ill-posed optimization problem into two parallel processes: semantic matching performed individually and reliability judgment for improving discrimination ability. Building on the idea of special tokens in language modeling representing sentence-level embeddings, we introduce a trusty token that enables distinguishing novel classes from known ones in prediction. To evaluate our approach, we conduct experiments on PASCAL VOC 2012, COCO-Stuff 164K and PASCAL Context. Our results show that TagCLIP improves the Intersection over Union (IoU) of unseen classes by 7.4\%, 1.7\% and 2.1\%, respectively, with negligible overheads. \rv{The code is available at \href{https://github.com/dvlab-research/TagCLIP}{here}}.
\end{abstract}

\begin{IEEEkeywords}
Computer Vision, Semantic Segmentation, Open-vocabulary learning, Vision language, Contrastive language-image pretraining
\end{IEEEkeywords}

%% file: sections/02_intro.tex
\IEEEPARstart{T}{he} \revise{long-term and challenging goal of deep learning \cite{mood, motcoder, moodv2, bal} is to approach human-level perception. Humans perceive scenes in a zero-shot manner, using multiple modalities such as vision, language, and sound. In line with this, researchers have proposed zero-shot algorithms for various computer vision tasks, including classification \cite{ovcls}, semantic segmentation \cite{ovseg, baseline, zegclip}, and object detection \cite{ovdet, ovdet_kd}.}

Although these approaches have achieved remarkable results \cite{fcn, unet, deeplab, deeplabv2}, deep learning models often fail to generalize to novel classes that were not seen during training. To improve the generalization ability of vision networks, previous researchers have leveraged recent advances in vision-language learning models, such as CLIP \cite{clip}, which learns rich multimodal features from a large-scale image-text dataset.

\revise{With the superior \textit{openness} provided by web-scale image-text pairs, vision-language models have been used in various ways \cite{baseline, ovseg, zegclip}.} Pioneers proposed two-stage approaches: \revise{first, generate class-agnostic proposals and then use pre-trained vision-language models like CLIP \cite{clip} to perform zero-shot classification. More recent approaches \cite{zegclip} boost both the performance and speed of zero-shot semantic segmentation by proposing one-stage approaches: they adopt a post-CLIP lightweight decoder that matches text prompts and image embeddings to generate segmentation maps.}

\revise{However, even the current state-of-the-art (SoTA) approach \cite{zegclip} struggles with pixels from novel classes and overfits them with seen semantics, such as \textit{cloud} with \textit{sky-other} and \textit{playing field} with \textit{dirt}.} To address this issue, we propose to disentangle the ill-posed optimization problem into two processes: one that performs semantic matching individually and another that determines prediction reliability and discriminates novel classes.

A straightforward solution is to apply outlier detectors \cite{mood} before feeding inputs to the downstream networks. However, an additional serial outlier detection stage inevitably doubles time and storage costs. In this work, we propose a novel framework, as shown in \cref{fig:framework}. Motivated by special tokens in language modeling that represent sentence-level embeddings \cite{feng2020language, takahashi2022unsupervised, bert}, we design an additional trusty token that denotes the prediction tendency and discriminates known and novel categories.

\begin{figure*}[t]
  \centering
  \includegraphics[width=\textwidth]{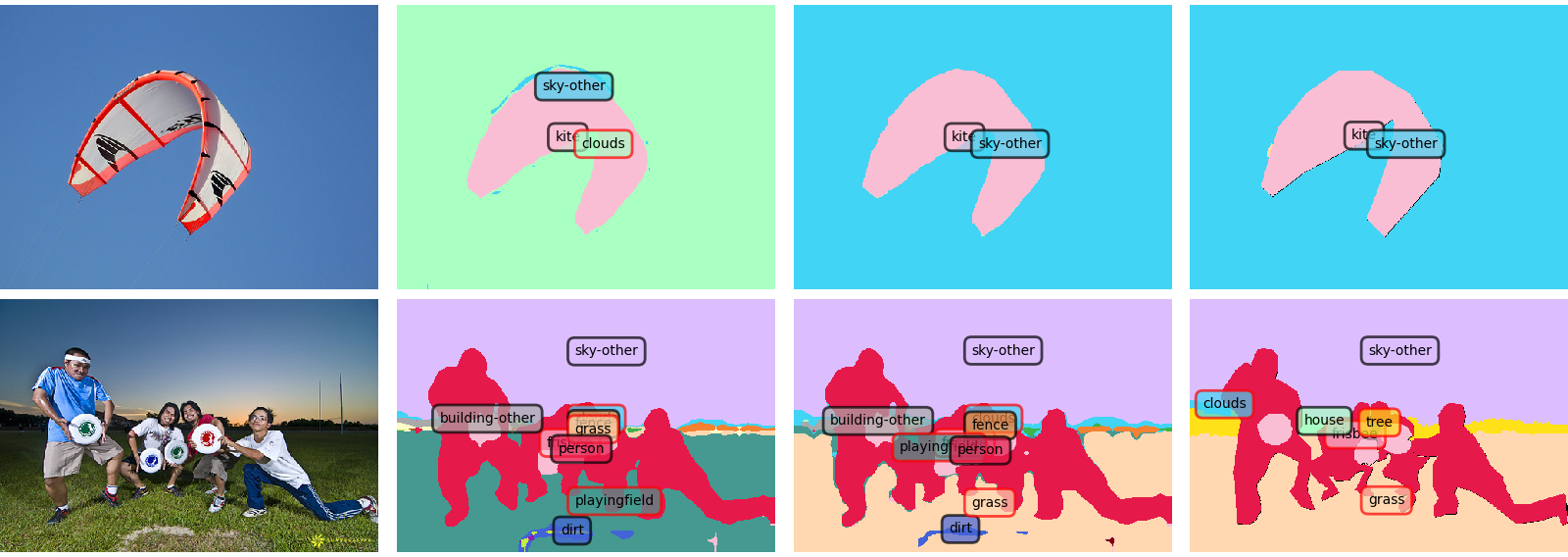}
  \caption{Visualization of segmentation results on COCO-Stuff 164K. Four columns from left to right represent (a) original testing images; (b) results of current SOTA \cite{zegclip}; (c) results of TagCLIP; (d) ground truth. The tags with borders in black and red denote seen and unseen classes separately.}
  \label{fig:vis}
\end{figure*}

We concatenate the trusty token with original classes and optimize it with our Trusty Learner, which performs trusty judgment with semantic matching in parallel. With almost no extra overhead, our TagCLIP (\textbf{T}rusty-\textbf{a}ware \textbf{g}uided CLIP) well discriminates unseen classes from known categories. \revise{As shown in \cref{fig:vis}, compared with the current SoTA, our proposed approach correctly separates hard unknown classes, including \textit{clouds}, \textit{playing field}, \textit{grass}, and others.} More visualization results are in \cref{sec:results}. 

We conducted extensive experiments to evaluate our proposed approach. Our method yields a significant improvement in Intersection over Union (IoU) of unseen classes by 7.4\% on PASCAL VOC 2012, by 1.7\% on COCO-Stuff 164K and by 2.1\% on PASCAL Context, as indicated in \cref{tab:results}. In the cross-dataset setting, our method enhances the performance of the state-of-the-art by 1.4\% from COCO-Stuff 164K to PASCAL Context, as shown in \cref{tab:cross_dataset}. \revise{We also present results trained with full labels in \cref{tab:supervision}, which demonstrate that our method improves the upper bound of zero-shot segmentation results by 1.9\% on PASCAL VOC 2012, by 0.4\% on COCO-Stuff 164K and by 0.9\% on PASCAL Context.}

In summary, the contributions of our method are:
\begin{enumerate}
\item We disentangle the ill-posed optimization into two parallel processes: one that judges reliability to improve discrimination and another that performs semantic matching individually.
\item We design a trusty token and optimize it with Trusty Learner, which discriminate novel classes from known categories with almost no extra overhead.
\item \revise{Our TagCLIP exhibits competitive performance in extensive zero-shot semantic segmentation tasks, including inductive settings, transductive settings, and cross-dataset tasks.}
\end{enumerate}

%% file: sections/03_relatex_works.tex
\subsection{Pretrained Vision Language Model}
Large-scale pretrained vision language models \cite{align, visualbert, clip, vlbert} that combine image representation and text embeddings have achieved impressive performance on multiple downstream tasks, including image retrieval \cite{liu2021image}, visual question answering \cite{jiang2022finetuning}, visual referring expression \cite{cris}, dense prediction \cite{denseclip}, and others. Among them, CLIP \cite{clip} is one of the most widely used vision-language models. CLIP is trained via contrastive learning on a billion-scale text-image dataset and has demonstrated its powerful generalization ability on various tasks~\cite{baseline, ovseg, zegclip}.

\subsection{\re{Semantic Segmentation}}
Semantic segmentation is an essential task in computer vision~\cite{fcn, zhang2023simple, ghiasi2022scaling, qin2023freeseg}. Pioneers addressed segmentation directly with per-pixel classification algorithms~\cite{fcn,segmenter,segformer, encnet, sert}, while followers improved them by decoupling mask generation with semantic classification~\cite{mask2former,maskformer,segvit}. Both principles have achieved significant progress when segmenting predefined closed categories. However, when inferring novel classes inaccessible from training, normal approaches perform poorly. Thus, zero-shot semantic segmentation has been a challenging task, where the key is to segment unseen categories via training on only known classes. Mainstream works~\cite{spnet, zs3, gagnet, sign, joint, strict} focus on improving the generalization ability from seen to novel categories.

\subsection{Zero-Shot Semantic Segmentation}
\revise{Inspired by the powerful generalization ability of CLIP~\cite{clip}, researchers have leveraged it for zero-shot semantic segmentation.} Early researchers proposed a two-stage paradigm~\cite{baseline, ovseg}: they first train a proposal generator and then utilize CLIP for pixel-level classification. The latest works~\cite{zegclip} simplify this process by proposing a one-stage approach that adds a lightweight transformer after CLIP as a decoder for segmentation. However, existing approaches still struggle with pixels from novel classes and overfit them with seen semantics. In our work, we proposed a novel framework called TagCLIP to improve discrimination ability, as shown in \cref{fig:framework}.

%% file: sections/04_methods.tex
\begin{figure*}[t]
    \centering
    \includegraphics[width=\textwidth,trim=85 125 205 100,clip]{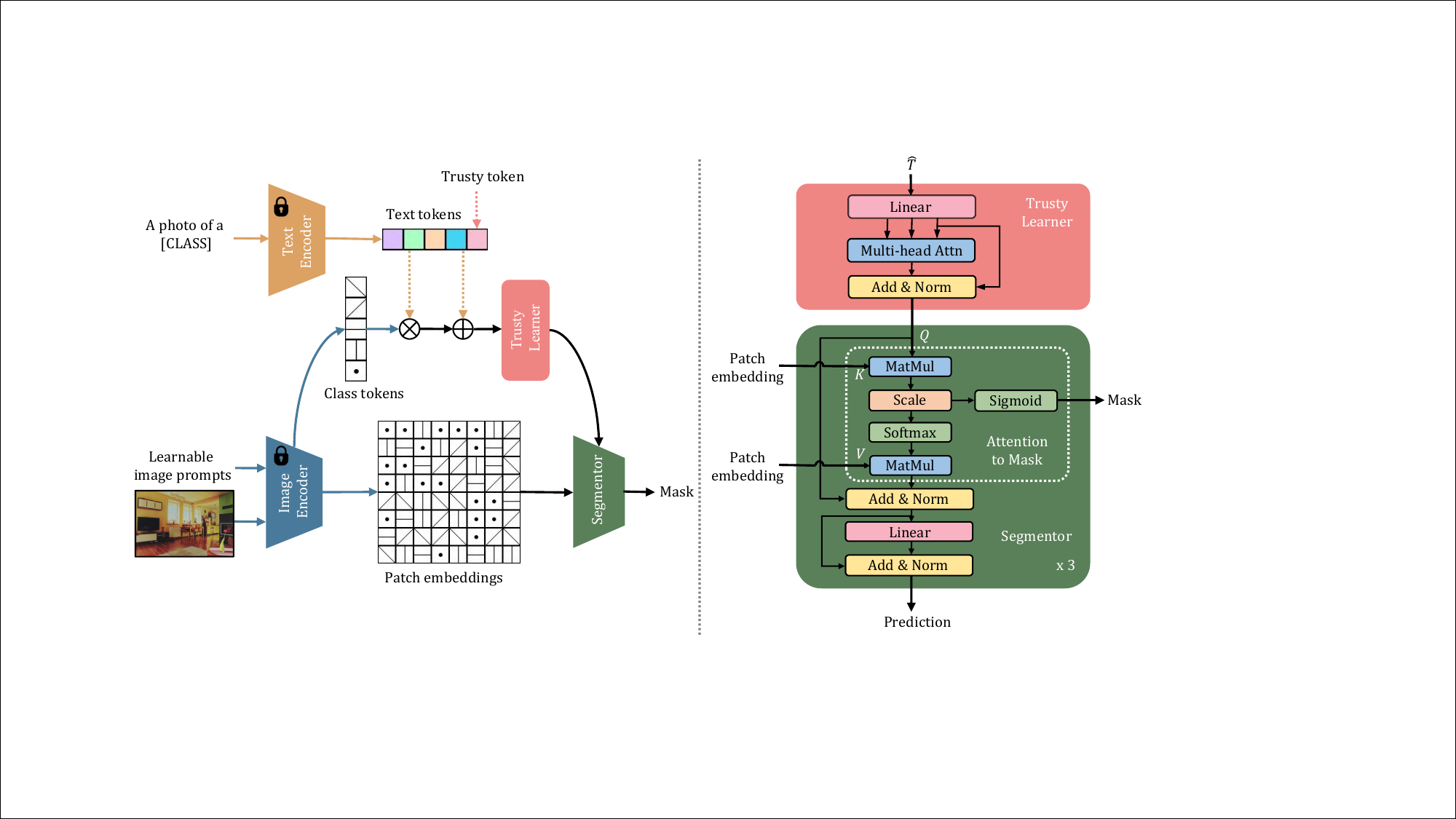}
    \caption{\revise{\textbf{Left: }The framework of our TagCLIP. First, we input images and text prompts into CLIP and concatenate a learnable trusty token with CLIP's text tokens. Then, we match the concatenated tokens with image tokens and input the output into our proposed Trusty Learner. Next, we perform a segmentor \cite{segvit}  to generate two maps: the trusty map and the raw mask. \re{$\otimes$ and $\oplus$ represent Hadamard product and concatenation.}
    \textbf{Right: }The detailed structure of Trusty Learner and Segmentor. The Trusty Learner contains a linear projection, a multi-head attention block with a shortcut, and a normalization layer. The segmentor contains three layers. Each layer constitutes an Attention-to-Mask block \cite{segvit} and a linear projection, both with shortcuts and normalization layers. }}
    \label{fig:framework}
\end{figure*}

\revise{Our method follows the literature of zero-shot semantic segmentation \cite{spnet}.} Its task is to train the model on only part of the classes but learned the capacity of segmenting both classes and novel categories. During training, the pixel annotations of unseen classes $C_U$ are masked and the seen part $C_S$ remains. During inference, the model is tested on the raw dataset with $C = C_U\cup C_S$.

In this section, we first introduce the framework of our proposed method (\cref{sec:framework}) and then propose the included techniques in training (\cref{sec:train}) and testing (\cref{sec:infer}) stages.

\subsection{Preliminary}
\label{sec:pre}

\revise{Recent work \cite{zegclip} has proposed an efficient one-stage zero-shot semantic segmentation pipeline.} In the one-stage framework, a vanilla light-weight transformer \cite{segvit} after the CLIP \cite{clip} matches class tokens and image embeddings extracted from the pre-trained CLIP. To be more specific, the process of the one-stage pipeline is as follows:


Firstly, we leverage deep prompt tuning to adapt CLIP \cite{clip} to our target dataset. Then, we extract CLIP's text tokens as $\mathbf{T} = [\mathbf{t}_1, \mathbf{t}_2, \dots, \mathbf{t}_C] \in \mathbb{R}^{C\times d}$ and \revise{class token} as $\mathbf{H} \in \mathbb{R}^{1\times d}$, where $C$ is the number of classes and $d$ is CLIP's feature dimension. Next, the input for \revise{segmentor} \cite{segvit} is:
\rv{
\begin{equation}\label{equ:relation}
    \mathbf{\hat{T}} = \phi_t({\rm concat}[\mathbf{T} \odot \mathbf{H}, \mathbf{T}]),
\end{equation}
where $\odot$ is the per-element Hadamard product and $\phi_t$ is a liner projection to align the dimension from $2d$ to $d$. The element-level representation of \cref{equ:relation} is:
\begin{equation}
    \mathbf{\hat{t}}_i = \phi_t({\rm concat}[\mathbf{t}_i\otimes\mathbf{H}, \mathbf{t}_i]), i=1,2,\dots,C,
\end{equation}
where $\otimes$ is the Hadamard product and $\mathbf{\hat{T}} = [\mathbf{\hat{t}}_1, \mathbf{\hat{t}}_2, \dots, \mathbf{\hat{t}}_C] \in \mathbb{R}^{C\times d}$.} Then linear projections $\phi$ are applied to generate $\mathbf{Q}(query)$, $\mathbf{K}(key)$ and $\mathbf{V}(value)$ as: 
\begin{equation}
\begin{aligned}
    \mathbf{Q} &= \phi_q(\mathbf{\hat{T}}) \in \mathbb{R}^{C\times d}, \\\mathbf{K} &= \phi_k(\mathbf{E}) \in \mathbb{R}^{N\times d}, \\\mathbf{V} &= \phi_v(\mathbf{E}) \in \mathbb{R}^{N\times d}.
\end{aligned}
\end{equation}
$\mathbf{E} =  [\mathbf{e}_1, \mathbf{e}_2, \dots, \mathbf{e}_N] \in \mathbb{R}^{N\times d}$ are the patch embeddings of the image encoder. Finally, the semantic masks \re{$\mathbf{M}_R$} are calculated by: 
\begin{equation}\label{equ:atm}
\re{\mathbf{M}_R} = {\rm Sigmoid}\left(\frac{\mathbf{Q}\mathbf{K}^T}{\sqrt{d}}\right) \in \mathbb{R}^{C \times N},
\end{equation}
where $\sqrt{d}$ is the scaling factor. The shape of \re{$\mathbf{M}_R$} is $C \times N$, and it can be further reshaped to \re{$C \times \frac{H}{P} \times \frac{W}{P}$}, where $P$ is the patch size.

\vspace{2mm}\noindent\re{\textbf{Final Segmentation Map.} The output class tokens of the segmentor is :
\begin{equation}
    \mathbf{\Theta}  = \text{Softmax}\left(\frac{\mathbf{Q}\mathbf{K}^T}{\sqrt{d}}\right)\mathbf{V} \in \rv{\mathbb{R}^{C\times d}}
\end{equation}
We employ a linear transformation $\phi_\pi$ on it to obtain class probability predictions. \rv{After that, the final segmentation map $\mathbf{\Pi}$ is computed as}:
\begin{equation}
    \mathbf{\Pi} = \phi_\pi (\mathbf{\Theta}) \cdot \mathbf{M}_R \in \mathbb{R}^{\frac{H}{P} \times \frac{W}{P}},
\end{equation}
Then $\mathbf{\Pi}$ is bilinearly upsampled to the original image size.
}

\begin{figure}[t]
    \centering
    \includegraphics[width=\textwidth,trim=80 85 75 55,clip]{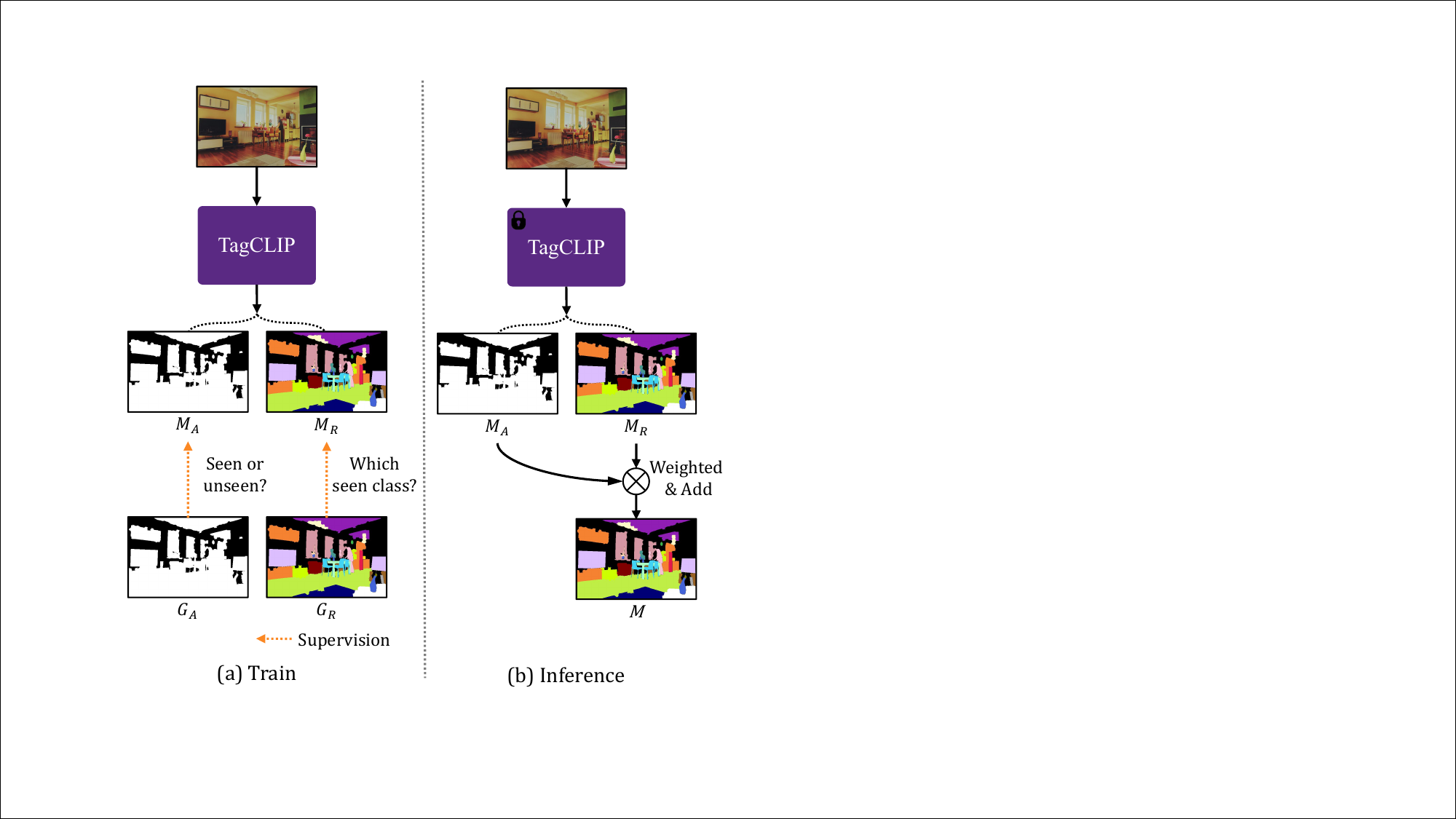}
    \caption{\revise{\textbf{Left: }During training, we propose a binary mask $\mathbf{G}_A$ for the supervision of $\mathbf{M}_A$ and utilize the ground truth $\mathbf{G}_R$ for the supervision of $\mathbf{M}_R$. \textbf{Right: }During inference, the raw semantic segmentation $\mathbf{M}_R$ is weighted by $\mathbf{M}_A$ to generate the final mask $\mathbf{M}$.}}
    \label{fig:seg}
\end{figure}

\begin{figure}[t]
    \centering
    \includegraphics[width=\linewidth]{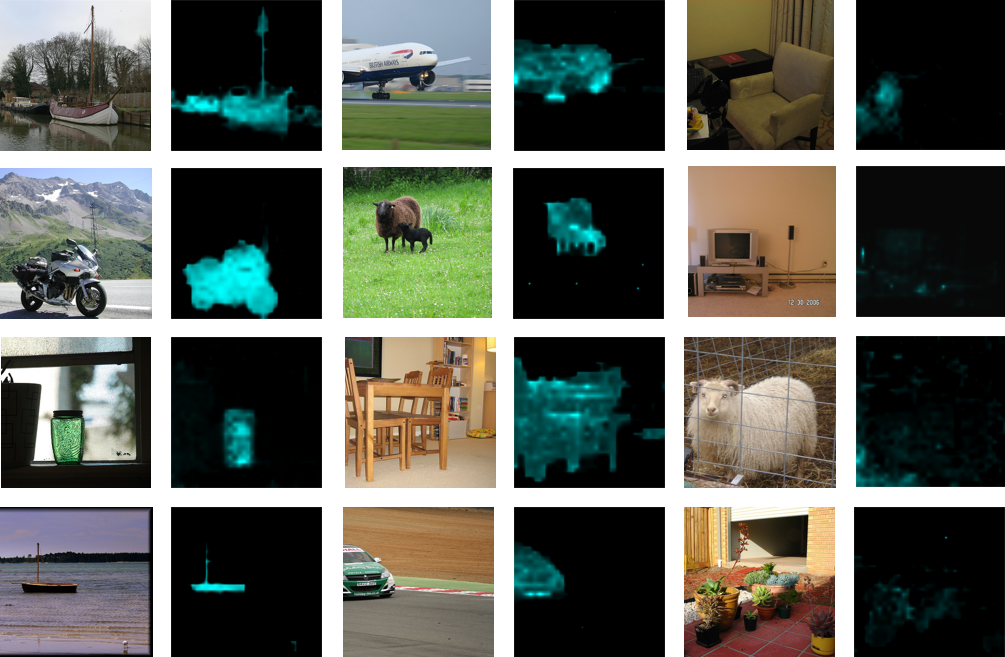}
    \caption{Visualization of the Trust Map. It exhibits a focused emphasis on seen classes (left 2 columns), namely airplane, bicycle, boat, car, cow, motorbike, dining table, bottle, etc. Concurrently, it effectively disregards unseen
    categories (right 1 column) such as potted plant, sheep, tv monitor, sofa, etc.}
    \label{fig:trusty_map}
\end{figure} 

\subsection{Our Framework}
\label{sec:framework}
In this section, we introduce our proposed framework, as shown in \cref{fig:framework}. It mainly contains the following steps:
\begin{enumerate}
    \item We perform the deep prompt tuning of CLIP.
    \item We concatenate a learnable trusty token with CLIP’s text tokens and match the concatenated tokens with CLIP's image tokens.
    \item We input the similarity matrix into a multi-head attention block.
    \item \revise{We perform a segmentor to generate two maps: the trusty map and the raw mask.}
\end{enumerate}


\subsubsection{Learnable Trusty Token.}
To address the issue of existing methods easily misclassifying input pixels from unseen classes, we propose to disentangle the ill-posed baseline into two processes: one performs semantic matching individually, and the other simultaneously judges prediction reliability to improve discrimination. We denote an additional trusty token as $\mathbf{t}_A \in \mathbb{R}^{1 \times d}$ to reflect the known and novel category prediction tendency. We concatenate $\mathbf{t}_A$ with the text tokens $\mathbf{T} = [\mathbf{t}_1, \mathbf{t}_2, \dots, \mathbf{t}_C] \in \mathbb{R}^{C \times d}$ to get $\mathbf{T}' = [\mathbf{t}_1, \mathbf{t}_2, \dots, \mathbf{t}_C, \mathbf{t}_A] \in \mathbb{R}^{(C+1) \times d}$, where $C$ is the number of classes and $d$ is the feature dimension of CLIP.

\subsubsection{Trusty Learner.}
\label{sec:rd}
\rv{Following \cite{zegclip}, we compute the per-element Hadamard product $\mathbf{T}' \in \mathbb{R}^{(C+1) \times d}$ with $\mathbf{H} \in \mathbb{R}^{1 \times d}$ and concatenate it with $\mathbf{T}'$: 
\begin{equation}\label{equ:hatt'}
    \mathbf{\hat{T}'} = \phi_t({\rm concat}[\mathbf{T'}\odot\mathbf{H}, \mathbf{T'}]),
\end{equation}
where $\phi_t$ is a liner projection to align the dimension from $2d$ to $d$}. Next, To capture the inter-relationship between CLIP's text tokens and inserted trusty tokens, we propose the Trusty Learner module. It constitutes a linear projection, a multi-head attention block with a shortcut, and a normalization layer, as shown in \cref{fig:framework}. The linear transformation firstly aligns the dimension from $2d$ to $d$. Then \re{$\mathbf{\hat{T}}’$ }is fed into the Trusty Learner and provides the output formulated by:
\re{
\begin{equation}
\mathbf{\tilde{T}} = {\rm Norm}({\rm Softmax}(\frac{\mathbf{\hat{T}}'\mathbf{\hat{T}}'^T}{\sqrt{d}})\mathbf{\hat{T}}'+\mathbf{\hat{T}}') \in \mathbb{R}^{(C+1)\times d},
\end{equation}
}
where \re{$\mathbf{\hat{T}}' = [\mathbf{\hat{t}}'_1, \mathbf{\hat{t}}'_2, \dots, \mathbf{\hat{t}}'_C, \mathbf{\hat{t}}'_A] \in \mathbb{R}^{(C+1) \times d}$} is the modified trusty token. Experiments in \cref{sec:ab} show that our proposed Trusty Learner is effective for instructing the trusty token $\mathbf{t}_A$.

\subsubsection{Trusty Map Generation.} \revise{Inspired by the research of \cite{segvit, zegclip}, we leverage the Attention-to-Mask (ATM) block to generate masks.} As shown in \cref{fig:seg}, the inputs are projected by $\phi$ to form query ($\mathbf{Q}$), key ($\mathbf{K}$), and values ($\mathbf{V}$) as follows:
\begin{equation}
\begin{aligned}
    \mathbf{Q} &= \mathbf{\tilde{T}} \in \mathbb{R}^{(C+1)\times d},  \\ \quad \mathbf{K} &= \phi_k(\mathbf{E}) \in \mathbb{R}^{N\times d},\\ \quad \mathbf{V} &= \phi_v(\mathbf{E}) \in \mathbb{R}^{N\times d},
\end{aligned}
\end{equation}
where $\mathbf{E} = [\mathbf{e}_1, \mathbf{e}_2, \dots, \mathbf{e}_N] \in \mathbb{R}^{N\times d}$ are the patch embeddings of CLIP's image encoder, $N$ is the number of patches, and $d$ is the feature dimension of CLIP. The semantic masks are calculated by:
\rv{
\begin{equation}\label{equ:atm}
\mathbf{M} = {\rm Sigmoid}\left(\frac{\mathbf{Q}\mathbf{K}^T}{\sqrt{d}}\right) \in \mathbb{R}^{(C+1) \times N},
\end{equation}
where $\sqrt{d}$ is the scaling factor. $\mathbf{M}$ can be further reshaped to $(C+1) \times \frac{H}{P} \times \frac{W}{P}$, where $P$ is the patch size. Note that we utilize ${\rm Sigmoid}$ as an activation function and the segmentation results of each class are independently generated. Therefore, $\mathbf{M}$ can be split into the trusty map $\mathbf{M}_A \in \mathbb{R}^{1\times \frac{H}{P} \times \frac{W}{P}}$ and raw semantic segmentation $\mathbf{M}_R \in \mathbb{R}^{C\times \frac{H}{P} \times \frac{W}{P}}$. We will introduce how to train and infer the maps next.
}

\subsection{Training Stage}
\label{sec:train}
In this section, we delve into the intricacies of our training strategy, focusing on two vital aspects: Trusty Labels and Losses. These components are integral to the success of our model, enhancing its ability to learn and generalize effectively. 

\subsubsection{Trusty Labels.} The training process is illustrated in \cref{fig:seg}(a). In order to supervise the learnable token $\mathbf{t}_A$, we create a pseudo map $\mathbf{G}_A$, with each pixel labeled 1 for seen class and 0 for the unseen class as:
\begin{equation}
\mathbf{G}_A(i, j) = \left\{\begin{matrix}
0 \textrm{,  if } \mathbf{G}_R(:, i, j) \in C_U \\
1 \textrm{,  if } \mathbf{G}_R(:, i, j) \in C_S,
\end{matrix}\right.
\end{equation}
where $C_U$ and $C_S$ are unseen and seen classes. $\mathbf{G}_R$ is the ground truth. In this way, although the model has no idea which unseen category the pixel is from, it discriminates unseen classes from known contexts. It has been experimentally demonstrated in \cref{tab:ab} that our design effectively improves the generalization ability to novel domains.

For a better interpretation of the trusty map, we visualize the attention map corresponding to the trusty token in \cref{fig:trusty_map}. It showcases the Trusty map, highlighting seen classes while effectively disregarding unseen categories. This visualization helps convey the ability of the trusty token to differentiate between seen and unseen classes, ultimately facilitating the model's discriminative power and enhancing overall performance.

\subsubsection{Losses.} For training the Trusty Learner, the loss $\mathcal{L}_A$ is defined as:
\begin{equation}
    \mathcal{L}_A = \mathcal{L}_{dice}(\mathbf{M}_A, \mathbf{G}_A),
\end{equation}
where $\mathcal{L}_{dice}$ is the dice loss \cite{dice, zegclip}, \re{$\mathbf{M}_A$ is the trusty map}. \revise{Integrated with the parallel branch that conducts semantic matching of zero-shot segmentation \cite{zegclip}, the overall loss during training is:}
\re{
\begin{equation}
\begin{aligned}
    \mathcal{L} &= \mathcal{L}_{mask} + \gamma\mathcal{L}_A \\
    &= \alpha\mathcal{L}_{focal}(\mathbf{M}_R, \mathbf{G}_R) \\ &+ \beta\mathcal{L}_{dice}(\mathbf{M}_R, \mathbf{G}_R) + \gamma\mathcal{L}_{dice}(\mathbf{M}_A, \mathbf{G}_A),
\end{aligned}
\end{equation}
}
where \re{$\{\mathcal{L}_{focal}, \mathcal{L}_{dice}\}$ are the focal loss \cite{focal}, and dice loss \cite{dice} with Sigmoid as activation function}. \revise{$\mathbf{M}_A$ and $\mathbf{M}_R$ are the trusty map and raw mask.} $\{\alpha, \beta, \gamma\}$ are weights of losses. We denote $\{\alpha, \beta, \gamma\}$ as $\{20, 1, 10\}$. More details are in \cref{sec:ab}.

\subsection{Inference Stage}
\label{sec:infer}
\vspace{2mm}\noindent\revise{\textbf{Mask.} The inference process is illustrated in \cref{fig:seg}(b). To make further use of the trusty map $\mathbf{M}_A$, we utilize $\mathbf{M}_A$ to weigh the confidence of the raw mask $\mathbf{M}_R$.} Each value $\mathbf{M}_A(i, j)$ in $\mathbf{M}_A$ represents the possibility of the pixel $(i, j)$ from seen classes and $(1-\mathbf{M}_A(i, j))$ represents that from novel categories. Consequently, we generate our final mask $\mathbf{M}$ by utilizing the trusty map $\mathbf{M}_A$ as weights of the raw mask $\mathbf{M}_R$:
\re{
\begin{equation}
\mathbf{M}(:, i, j) = \begin{cases}
\mathbf{M}_R(:, i, j) (1-\mathbf{M}_A(i, j)), & \text{if } \mathbf{M}_A(i, j) \in C_U, \\
\mathbf{M}_R(:, i, j) \mathbf{M}_A(i, j), & \text{if } \mathbf{M}_A(i, j) \in C_S,
\end{cases}
\end{equation}
}
where $C_U$ and $C_S$ are unseen and seen classes. \revise{Experiments in \cref{sec:ab} demonstrate a better performance of the weighted map $\mathbf{M}$ compared with the raw mask $\mathbf{M}_R$.}

\begin{table*}[t]
\small
  \caption{Detailed experimental configuration. 100/10 represents 100 for COCO and 10 for VOC.}
  \setlength{\tabcolsep}{3.5mm}
  \label{sample-table}
  \centering
  \begin{tabular}{c|cc|c|cc}
  \toprule
\multicolumn{2}{c}{config}  & value  & \multicolumn{2}{c}{config}& value  \\
\midrule
\multirow{8}{*}{backbone} & num token  & 100/10 & \multirow{6}{*}{learning rate config} & policy & poly  \\
  & patch size & 16 & & power  & 0.9 \\
  & width  & 768  & & minimal learning rate  & $1\times 10^{-6}$ \\
  & output dim & 512  & & warmup & linear  \\
  & drop path rate & 0.1  & & warmup iters & 1500  \\
  & layers & 12 & & warmup ratio & $1\times 10^{-6}$ \\ \cline{4-6} 
  & input resolution & 512  & \multirow{8}{*}{optimizer}  & type & AdamW \\
  & prompt dimension & 716  & & learning rate  & $2\times 10^{-5}$ \\ \cline{1-3}
\multirow{5}{*}{text encoder} & context length & 77 & & weight decay & 0.01  \\
  & embed dimension  & 512  & & backbone lr multiplier & 10.0  \\
  & transformer width  & 512  & & head & 10.0  \\
  & transformer heads  & 8  & & text encoder & 0.0 \\
  & transformer layers & 12 & & norm & 0.0 \\ \cline{1-3}
\multirow{4}{*}{segmentor}  & input resolution & 512  & & ln & 0.0 \\ \cline{4-6} 
  & input channels & 512  & \multirow{3}{*}{loss} & $\alpha$ & 20.0  \\
  & number of layers & 3  & & $\beta$  & 1.0 \\
  & number of heads  & 8  & & $\gamma$ & 10.0 \\
  \bottomrule
  \end{tabular}
  \label{tab:cfg}
\end{table*}

\begin{table*}[t]
\small
  \caption{Ablation experiment results on PASCAL VOC 2012, including results (a) of one-stage baseline, (b) with trusty token, (c) with attention block, and (d) with weighted map (TagCLIP). \re{$\mathbf{H}$: class token; $\mathbf{T}'$: concatenation of text tokens and trusty token; $\mathbf{T}'\odot \mathbf{H}$: per-element Hadamard product; $[\mathbf{T}', \mathbf{H}]$: concatenation along the dimension. $\mathbf{\hat{T}}'$: defined in \cref{equ:hatt'}.} Enabled: whether this architecture is included in the framework.}
  \label{tab:ab}
  \setlength{\tabcolsep}{3.2mm}
  \centering
  \begin{tabular}{c|c|cccc|c|cccc}
    \toprule
\multirow{2}{*}{\textit{}}     & Trusty   & \multicolumn{4}{c|}{Trusty Learner} & Weighted & \multicolumn{4}{c}{Metrics}     \\
 & token    & Enabled?    & $\mathbf{Q}$  & $\mathbf{K}$  & $\mathbf{V}$  & map & pAcc     & mIoU(S)  & mIoU(U)  & hIoU     \\
 \midrule
(\textit{a})    & {\color{gray}\ding{56}} & {\color{gray}\ding{56}} & -   & -   & -   & {\color{gray}\ding{56}} & \rv{95.0} & \rv{92.6} &  \rv{75.5}   & \rv{83.2}  \\ 
 \midrule
(\textit{b})    & \ding{52}     & {\color{gray}\ding{56}} & -   & -   & -   & {\color{gray}\ding{56}} & {\color{gray}90.4} & {\color{gray}91.6} & {\color{gray}43.0} & {\color{gray}58.6} \\
 \midrule
\multirow{13}{*}{(\textit{c})} & \ding{52}     & \ding{52}     & $\mathbf{H}$  & $\mathbf{H}$  & \re{$\mathbf{T}'$}  & {\color{gray}\ding{56}} & {\color{gray}87.4} & {\color{gray}89.9} & {\color{gray}35.0} & {\color{gray}50.4} \\
 & \ding{52}     & \ding{52}     & \re{$\mathbf{T}'$}  & $\mathbf{H}$  & $\mathbf{H}$  & {\color{gray}\ding{56}} & {\color{gray}84.5} & {\color{gray}85.3} & {\color{gray}30.3} & {\color{gray}44.7} \\
 & \ding{52}     & \ding{52}     & \re{$\mathbf{T}'$}  & $\mathbf{H}$  & \re{$\mathbf{T}'$}  & {\color{gray}\ding{56}} & {\color{gray}82.8} & {\color{gray}90.2} & {\color{gray}13.5} & {\color{gray}23.4} \\
 & \ding{52}     & \ding{52}     & \re{$\mathbf{T}'$}  & \re{$\mathbf{T}'$}  & $\mathbf{H}$  & {\color{gray}\ding{56}} & {\color{gray}84.5} & {\color{gray}84.3} & {\color{gray}30.8} & {\color{gray}45.1} \\
 & \ding{52}     & \ding{52}     & \re{$\mathbf{T}'$}  & \re{$\mathbf{T}'$}  & \re{$\mathbf{T}'\odot \mathbf{H}$} & {\color{gray}\ding{56}} & {\color{gray}82.0}   & {\color{gray}89.2} & {\color{gray}14.7} & {\color{gray}25.2} \\
 & \ding{52}     & \ding{52}     & \re{$[\mathbf{T}', \mathbf{H}]$}   & \re{$[\mathbf{T}', \mathbf{H}]$}   & \re{$[\mathbf{T}', \mathbf{H}]$}   & {\color{gray}\ding{56}} & {\color{gray}88.5} & {\color{gray}90.1} & {\color{gray}41.5} & {\color{gray}56.8} \\
 & \ding{52}     & \ding{52}     & \re{$\mathbf{T}'\odot \mathbf{H}$} & $\mathbf{H}$  & $\mathbf{H}$  & {\color{gray}\ding{56}} & {\color{gray}85.4} & {\color{gray}90.4} & {\color{gray}23.5} & {\color{gray}37.3} \\
 & \ding{52}     & \ding{52}     & \re{$\mathbf{T}'\odot \mathbf{H}$} & \re{$\mathbf{T}'$}  & \re{$\mathbf{T}'$}  & {\color{gray}\ding{56}} & {\color{gray}86.5} & {\color{gray}90.4} & {\color{gray}30.0} & {\color{gray}45.0} \\
 & \ding{52}     & \ding{52}     & \re{$\mathbf{T}'\odot \mathbf{H}$} & \re{$\mathbf{T}'$}  & \re{$\mathbf{T}'\odot \mathbf{H}$} & {\color{gray}\ding{56}} & {\color{gray}82.5} & {\color{gray}89.3} & {\color{gray}13.3} & {\color{gray}23.2} \\
 & \ding{52}     & \ding{52}     & \re{$\mathbf{T}'\odot \mathbf{H}$} & \re{$\mathbf{T}'\odot \mathbf{H}$} & $\mathbf{H}$  & {\color{gray}\ding{56}} & {\color{gray}85.2} & {\color{gray}90.6} & {\color{gray}23.6} & {\color{gray}37.5} \\
 & \ding{52}     & \ding{52}     & \re{$\mathbf{T}'\odot \mathbf{H}$} & \re{$\mathbf{T}'\odot \mathbf{H}$} & \re{$\mathbf{T}'$}  & {\color{gray}\ding{56}} & {\color{gray}86.2} & {\color{gray}86.6} & {\color{gray}38.0} & {\color{gray}52.8} \\
 & \ding{52}     & \ding{52}     & \re{$\mathbf{T}'\odot \mathbf{H}$} & \re{$\mathbf{T}'\odot \mathbf{H}$} & \re{$\mathbf{T}'\odot \mathbf{H}$} & {\color{gray}\ding{56}} & {\color{gray}83.3} & {\color{gray}89.9} & {\color{gray}25.1} & {\color{gray}39.2} \\
 & \ding{52}     & \ding{52}     &\re{$\mathbf{\hat{T}}’$ }   & \re{$\mathbf{\hat{T}}’$ }   & \re{$\mathbf{\hat{T}}’$ }   & {\color{gray}\ding{56}} & \textbf{96.2} & \textbf{93.6} & \textbf{84.9} & \textbf{89.0} \\
  \midrule
(\textit{d})    & \ding{52}     & \ding{52}     & \re{$\mathbf{\hat{T}}’$ }  & \re{$\mathbf{\hat{T}}’$ }   & \re{$\mathbf{\hat{T}}’$ }   & \ding{52}     & \textbf{96.1} & \textbf{93.5} & \textbf{85.2} & \textbf{89.2} \\
    \bottomrule
  \end{tabular}
\end{table*}

\begin{table}[t]
\small
  \caption{\revise{The unseen class names of PASCAL VOC 2012, COCO-Stuff 164K and PASCAL Context}}
  \setlength{\tabcolsep}{5mm}
  \label{sample-table}
  \centering
  \begin{tabular}{cc}
  \toprule
Dataset & The name of unseen classes  \\
\midrule
VOC & \textit{pottedplant, sheep, sofa, train, tvmonitor} \\
\midrule
\multirow{3}{*}{COCO} & \textit{cow, giraffe, suitcase, frisbee, skateboard}  \\
 & \textit{playingfield, river, road, tree, wall concrete}  \\
  & \textit{cardboard, clouds, gras, carrot, scissors}  \\
\midrule
\multirow{2}{*}{\revise{Context}} & \textit{cow, motorbike, sofa, cat, boat}  \\
 & \textit{fenc, bird, tv monitor, keyboard, aeroplane} \\
  \bottomrule
  \end{tabular}
  \label{tab:unseen_class_names}
\end{table}

\begin{figure}[t]
   \centering
   \includegraphics[width=\linewidth,trim=185 255 405 150,clip]{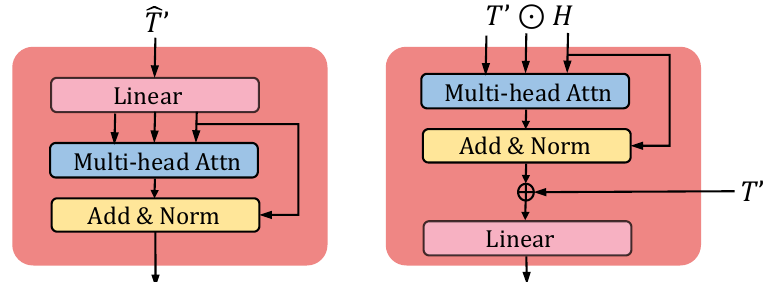}
   \caption{\rv{Examples of two Trusty Learner structures. \textbf{Left }concatenates $\mathbf{T'}\odot \mathbf{H}$ and $\mathbf{T'}$ before inputting them into the Multi-head attention. \textbf{Right }first inputs $\mathbf{T'}\odot \mathbf{H}$ into the Multi-head attention and then concatenates it with $\mathbf{T'}$. $\oplus$: dimension concatenation. $\odot$: per-element Hadamard product. $\mathbf{\hat{T'}}$: defined in \cref{equ:hatt'}. }}
   \label{fig:ab_structure}
\end{figure}

\begin{figure}[t]
   \centering
   \includegraphics[width=0.8\linewidth,trim=0 0 0 0,clip]
   {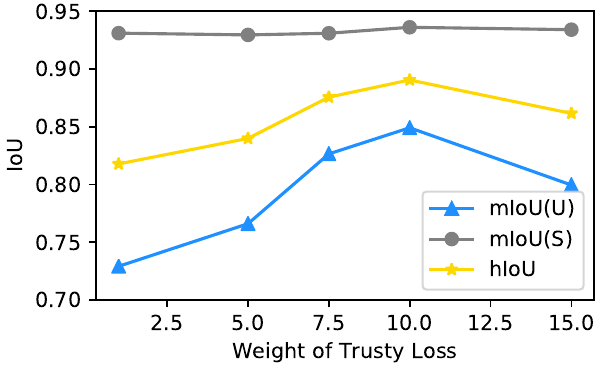}
   \caption{Effect of the weight of trusty loss.}
   \label{fig:loss_weight}
\end{figure}

\begin{table}[t]
\centering
\caption{\textbf{Initialization of Trusty token}. Employing an average of three testing iterations, evaluations were conducted on Pascal VOC. \revise{Bold indicates best performance and gray background indicates ours.}}
\setlength{\tabcolsep}{3.3mm}
\small
\begin{tabular}{c|cccc}
\toprule
Initial & pAcc      & mIoU(S)   & mIoU(U) & hIoU      \\
\midrule
Rand    & 95.6      & 93.0      & 81.2    & 86.7      \\
\rowcolor{gray!20} \revise{Zero} & \textbf{\revise{96.1}} & \textbf{\revise{93.5}} & \textbf{\revise{85.2}} & \textbf{\revise{89.2}}\\
\bottomrule
\end{tabular}
\label{tab:init}
\end{table}

\begin{table*}[t]
\small
  \caption{\textbf{Inductive task}: Comparison with previous methods. \revise{Bold indicates best performance and gray background indicates ours.} \re{`-' represents the method does not include a text encoder}. \rv{w2v: word2vec, ft: fasttext.}}
  \label{tab:results}
  \setlength{\tabcolsep}{0.8mm}
  \centering
  \begin{tabular}{c|c|cccc|cccc|cccc}
    \toprule
\multirow{2}{*}{Methods} & \revise{Text} & \multicolumn{4}{c|}{PASCAL VOC 2012} & \multicolumn{4}{c|}{COCO-stuff 164K} & \multicolumn{4}{c}{PASCAL Context} \\
 & \revise{Encoder} & pAcc & mIoU(S) & mIoU(U) & hIoU & pAcc & mIoU(S) & mIoU(U) & hIoU & pAcc & mIoU(S) & mIoU(U) & hIoU \\
 \midrule
SPNet \cite{spnet} & \rv{w2v \cite{word2vec}, ft \cite{fasttext}} & - & 78.0 & 15.6 & 26.1 & - & 35.2 & 8.7 & 14.0 & - & - & - & - \\
ZS3 \cite{zs3} & \rv{w2v \cite{word2vec}} & - & 77.3 & 17.7 & 28.7 & - & 34.7 & 9.5 & 15.0 & 52.8 & 20.8 & 12.7 & 15.8 \\
\re{CAGNet} \cite{gagnet} & \rv{w2v \cite{word2vec}} & 80.7 & 78.4 & 26.6 & 39.7 & 56.6 & 33.5 & 12.2 & 18.2 & - & 24.1 & 18.5 & 21.2 \\
SIGN \cite{sign} & \rv{w2v \cite{word2vec}} & - & 75.4 & 28.9 & 41.7 & - & 32.3 & 15.5 & 20.9 & - & - & - & - \\
Joint \cite{joint} & \rv{w2v \cite{word2vec}} & - & 77.7 & 32.5 & 45.9 & - & - & - & - & - & 33.0 & 14.9 & 20.5 \\
ZegFormer \cite{zegformer} & \re{CLIP \cite{clip}} & - & 86.4 & 63.6 & 73.3 & - & 36.6 & 33.2 & 34.8 & - & - & - & - \\
zsseg \cite{baseline} & \re{CLIP \cite{clip}} & 90.0 & 83.5 & 72.5 & 77.5 & 60.3 & 39.3 & 36.3 & 37.8 & - & - & - & - \\
\revise{FreeSeg \cite{qin2023freeseg}} & \re{CLIP \cite{clip}} & - & \revise{91.8} & \revise{82.6} & \revise{86.9} & - & - & - & - & - & - & - & - \\
\revise{ZegCLIP} \cite{zegclip}  & \re{CLIP \cite{clip}} & 94.6 & 91.9 & 77.8 & 84.3 & 62.0 & 40.2 & 41.4 & 40.8 & 76.2 & 46.0 & 54.6 & 49.9 \\
\rowcolor{gray!20} \revise{TagCLIP}   & \re{CLIP \cite{clip}}  & \textbf{\revise{96.1}}& \textbf{\revise{93.5}} & \textbf{\revise{85.2}} & \textbf{\revise{89.2}} & \textbf{\revise{63.3}} & \textbf{\revise{40.7}} & \textbf{\revise{43.1}} & \textbf{\revise{41.9}} & \textbf{\revise{76.8}} & \textbf{\revise{48.1}} & \textbf{\revise{56.7}} & \textbf{\revise{52.1}} \\
    \bottomrule
  \end{tabular}
\end{table*}

\begin{figure*}[t]
  \centering
 \includegraphics[width=0.8\textwidth,trim=0 0 0 0,clip]
 {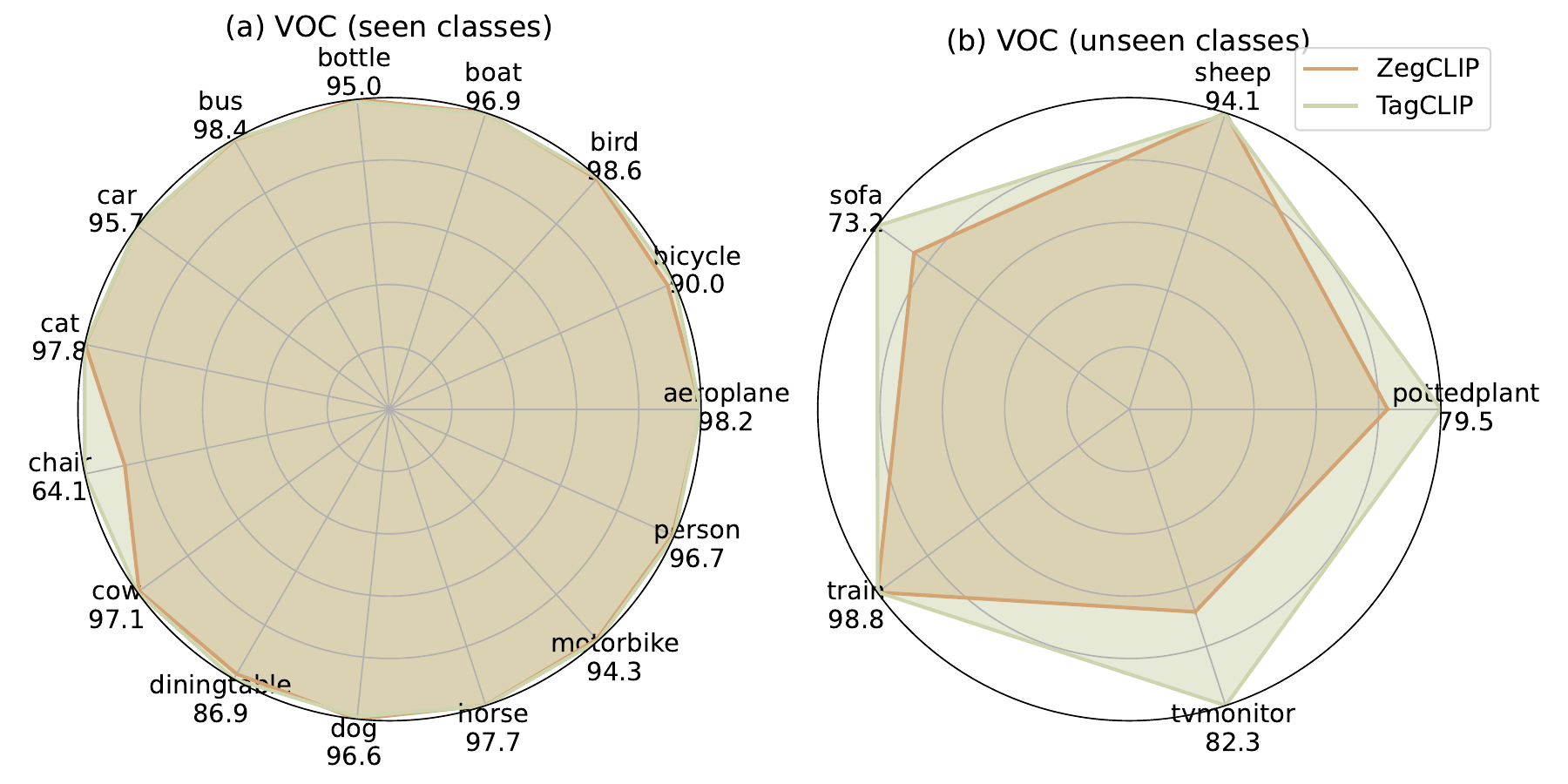}
  \caption{\revise{Detailed performance of (a) seen classes and (b) unseen classes of PASCAL VOC 2012.}}
  \label{fig:1_class}
\end{figure*}

\begin{figure*}[p]
  \centering
  \includegraphics[width=\textwidth]{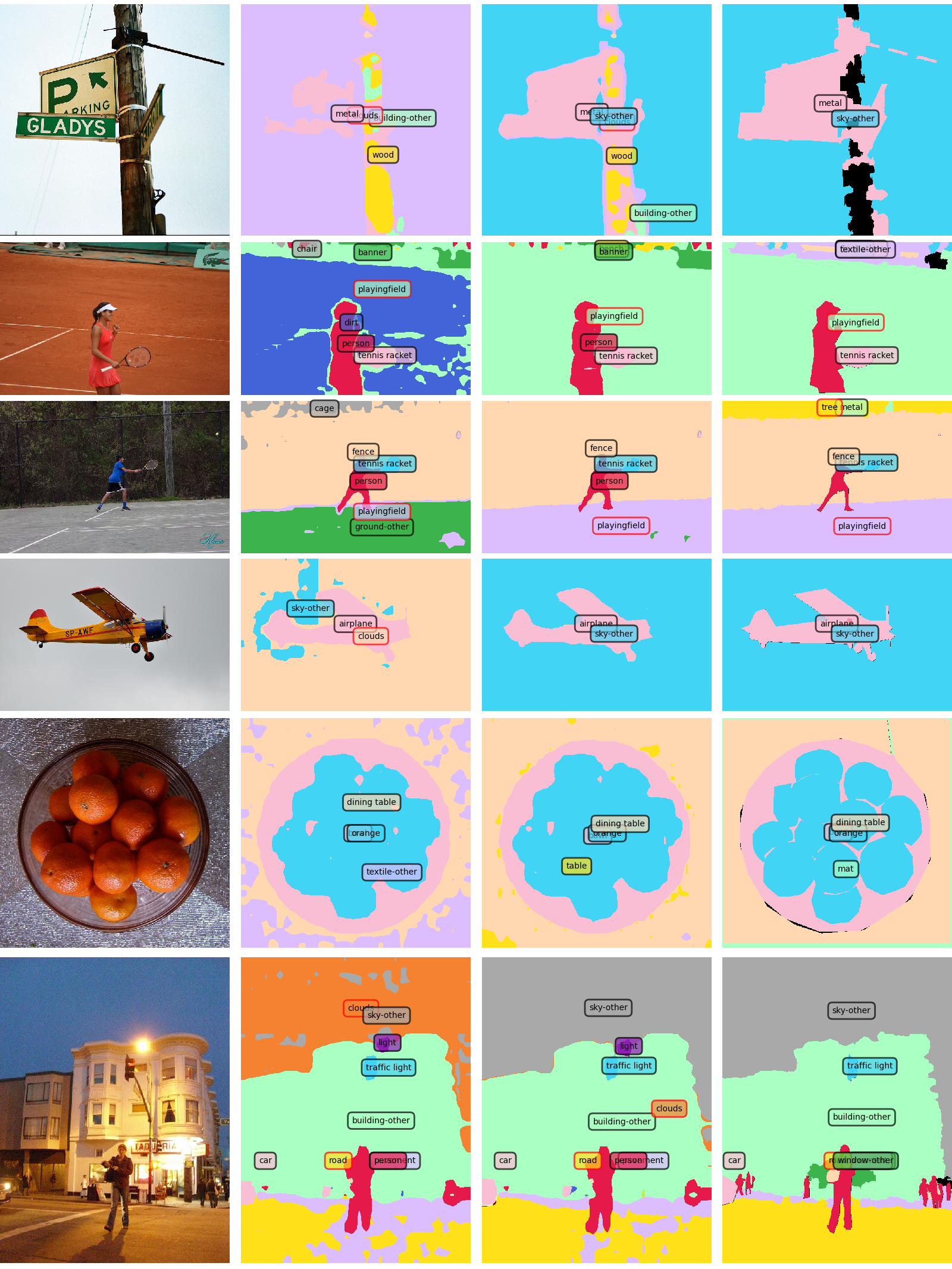}
  \caption{\revise{Visualization of segmentation results on COCO-Stuff 164K. Four columns from left to right represent: (a) original testing images; (b) results of current SoTA \cite{zegclip}; (c) results of TagCLIP; (d) ground truth. The tags with borders in black and red denote seen and unseen classes separately.}}
  \label{fig:vis_exp}
\end{figure*}

\begin{table}[t]
\centering
\caption{\textbf{Parameters and FPS}. Comparison of inference speed in terms of Frames Per Second (FPS) and learnable parameters. \revise{Bold indicates best performance and gray background indicates ours.}}
\setlength{\tabcolsep}{4.3mm}
\small
\begin{tabular}{c|c|ccc}
\toprule
\multirow{2}{*}{Method} & \multirow{2}{*}{Params  (M)}  & \multicolumn{2}{c}{FPS}         \\
         &         & VOC    & COCO   \\
\midrule
ZegCLIP \cite{zegclip} & 13.8   & 9.0    & 6.7    \\
\rowcolor{gray!20}\revise{TagCLIP}  & \textbf{\revise{11.7}}\tiny{\color{teal}(-15\%)} & \textbf{\revise{13.3}}\tiny{\color{teal}(+47\%)} & \textbf{\revise{10.1}}\tiny{\color{teal}(+51\%)} \\
\bottomrule
\end{tabular}
\label{tab:param}
\end{table}

\begin{table}[t]
   \small
   \centering
   \caption{Generalization performance from COCO-Stuff 164K to PASCAL Context. \revise{Bold indicates best performance and gray background indicates ours.}}
   \setlength{\tabcolsep}{5.7mm}
   \begin{tabular}{c|ccccc}
   \toprule
   Method     & pAcc & mIoU & mAcc \\ 
   \midrule
   Zegformer \cite{zegformer} & 42.3 & 29.3 & 56.6 \\
   ZegCLIP \cite{zegclip}     & 60.9 & 41.2 & 68.4 \\
\rowcolor{gray!20} \revise{TagCLIP}  & \textbf{\revise{63.5}} & \textbf{\revise{42.6}} & \textbf{\revise{70.1}} \\
   \bottomrule
   \end{tabular}
   \label{tab:cross_dataset}
\end{table}

\begin{table*}[t]
\small
  \caption{\textbf{Transductive task}: Comparison with previous methods. \textit{ST}: Self-Training. \revise{Bold indicates best performance and gray background indicates ours.} \re{`-' represents the method does not include a text encoder}. \rv{w2v: word2vec, ft: fasttext.}}
  \label{tab:transductive}
  \setlength{\tabcolsep}{0.6mm}
  \centering
  \begin{tabular}{c|c|cccc|cccc|cccc}
    \toprule
\multirow{2}{*}{Methods} & \revise{Text} & \multicolumn{4}{c|}{PASCAL VOC 2012} & \multicolumn{4}{c|}{COCO-stuff 164K} & \multicolumn{4}{c}{PASCAL Context} \\
 & \revise{Encoder}  & pAcc & mIoU(S) & mIoU(U) & hIoU & pAcc & mIoU(S) & mIoU(U) & hIoU & pAcc & mIoU(S) & mIoU(U) & hIoU \\
 \midrule
SPNet + ST \cite{spnet}  & \rv{w2v \cite{word2vec}, ft \cite{fasttext}}  & - & 77.8 & 25.8 & 38.8 & - & 34.6 & 26.9 & 30.3 & - & - & - & - \\
ZS5 \cite{zs3} & \rv{w2v \cite{word2vec}} & - & 78.0 & 21.2 & 33.3 & - & 34.9 & 10.6 & 16.2 & 49.5 & 27.0 & 20.7 & 23.4 \\
\re{CAGNet} + ST \cite{gagnet} & \rv{w2v \cite{word2vec}} & 81.6 & 78.6 & 30.3 & 43.7 & 56.8 & 35.6 & 13.4 & 19.5 & - & - & - & - \\
STRICT \cite{strict}  & \rv{w2v \cite{word2vec}} & - & 82.7 & 35.6 & 49.8 & - & 35.3 & 30.3 & 34.8 & - & - & - & - \\
zsseg + ST \cite{baseline} & \re{CLIP \cite{clip}} & 88.7 & 79.2 & 78.1 & 79.3 & 63.8 & 39.6 & 43.6 & 41.5 & - & - & - & - \\
MaskCLIP+ \cite{denseclip} & \re{CLIP \cite{clip}} & - & 88.8 & 86.1 & 87.4 & - & 38.1 & 54.7 & 45.0 & - & 44.4 & 66.7 & 53.3 \\
\revise{ZegCLIP} +ST \cite{zegclip} & \re{CLIP \cite{clip}} & 96.2 & 92.3 & 89.9 & 91.1 & 69.2 & 40.6 & 59.9 & 48.4 & 77.3 & 46.8 & 68.5 & 55.6 \\
\rowcolor{gray!20}\revise{TagCLIP + ST} & \re{CLIP \cite{clip}} & \textbf{\revise{97.2}} & \textbf{\revise{94.3}} & \textbf{\revise{92.7}} & \textbf{\revise{93.5}} & \textbf{\revise{69.4}} & \textbf{\revise{40.8}} & \textbf{\revise{60.0}} & \textbf{\revise{48.6}} & \textbf{\revise{78.0}} & \textbf{\revise{49.2}} & \textbf{\revise{65.8}} & \textbf{\revise{55.9}}  \\
    \bottomrule
  \end{tabular}
\end{table*}

\begin{table*}[t]
\small
  \caption{\textbf{Supervision task}: Comparison with previous methods. \revise{Bold indicates best performance and gray background indicates ours.}}
  \label{tab:supervision}
  \setlength{\tabcolsep}{2.3mm}
  \centering
  \begin{tabular}{c|c|ccc|ccc|ccc}
    \toprule
\multirow{2}{*}{Methods} & \revise{Text} & \multicolumn{3}{c|}{PASCAL VOC 2012} & \multicolumn{3}{c|}{COCO-stuff 164K} & \multicolumn{3}{c}{PASCAL Context} \\
 & \revise{Encoder} & pAcc & mIoU(S) & mIoU(U) & pAcc & mIoU(S) & mIoU(U) & pAcc & mIoU(S) & mIoU(U)\\
\midrule
\revise{ZegCLIP} \cite{zegclip} & \re{CLIP \cite{clip}} & 96.3 & 92.0 & - & 69.9 & 42.6 & - & 77.5 & 51.1 & -  \\
\rowcolor{gray!20}\revise{TagCLIP} & \re{CLIP \cite{clip}} & \textbf{\revise{96.8}} & \textbf{\revise{93.0}} & - & \textbf{\revise{69.9}} & \textbf{\revise{43.2}} & - & \textbf{\revise{77.9}} & \textbf{\revise{52.0}} & - \\
    \bottomrule
  \end{tabular}
\end{table*}

\begin{table}[t]
\small
  \caption{\re{Performance of COCO-Stuff-164K in the setting which can't differentiate between novel classes and the background. Bold indicates best performance and gray background indicates ours.}}
  \label{tab:supervision}
   \setlength{\tabcolsep}{3.5mm}
   \begin{tabular}{c|ccccc}
   \toprule
   Method     & pAcc & mIoU(S) & mIoU(U) & hIoU \\ 
   \midrule
   ZegCLIP \cite{zegclip}     & 47.2 & 29.9 & 30.4 & 30.2\\
\rowcolor{gray!20} TagCLIP & \textbf{50.7} & \textbf{30.0} & \textbf{36.0} & \textbf{32.7} \\
   \bottomrule
   \end{tabular}
   \label{tab:bg}
\end{table}

%% file: sections/05_experiment.tex
 In this extensive experimental section, we provide a comprehensive assessment of our approach's performance. We introduce the datasets used (\cref{sec:datasets}), present the evaluation metrics employed (\cref{sec:metrics}), and outline our experimental setup (\cref{sec:setting}). In \cref{sec:ab}, we conduct a detailed ablation analysis to dissect the contributions of key components. Finally in \cref{sec:results}, we present the results, demonstrating the effectiveness and versatility of our method across various datasets and settings. These experiments collectively affirm the robustness and performance of our proposed approach.

\subsection{Datasets} 
\label{sec:datasets}
The experiments are conducted on the following public benchmark datasets: 
\begin{enumerate}
    \item \textbf{PASCAL VOC 2012}, which contains 20 classes with 10,582 augmented training and 1,449 testing images. We divide the dataset into 15 seen classes and 5 unseen classes by ignoring the "background" category.
    \item \textbf{COCO-Stuff 164K}, which includes 171 categories with 118,287 training and 5,000 testing images. We use 156 and 15 classes as the seen and unseen parts.
    \item \textbf{PASCAL Context}, which contains 60 classes with 4,996 training and 5,104 testing images. 
\end{enumerate} 
We follow the same unseen classes with previous works \cite{zegformer, gagnet, baseline, denseclip, zegclip}, as shown in \cref{tab:unseen_class_names}.

\subsection{Evaluation Metrics} 
 \label{sec:metrics}
 Following previous works \cite{baseline, zegclip}, we evaluate the performance of our model using two metrics: pixel-wise classification accuracy (pAcc) and mean of class-wise Intersection over Union (mIoU). We denote mIoU on seen and unseen classes as mIoU(S) and mIoU(U), respectively. In addition, we measure the harmonic mean IoU (hIoU) among seen and unseen classes, which is defined as
\begin{equation}
hIoU = \frac{2\times mIoU(S)\times mIoU(U)}{mIoU(S)+ mIoU(U)}
\end{equation}

\subsection{Experimental Setting}
\label{sec:setting}
All experiments are based on a pre-trained CLIP ViT-B/16 model. The results of previous works are from \cite{zegclip}. We conduct our experiment on 8 GPUs, with an input resolution set to $512\times512$ and a batch size of 16. To ensure fair comparisons with previous works \cite{zegclip}, in the \textit{inductive} zero-shot learning setting, we train the model with 20K iterations for PASCAL VOC 2012 and 80K for COCO-Stuff 164K. In the \textit{transductive} setting, we train our proposed model with 10K iterations for PASCAL VOC 2012 and 40K for COCO-Stuff 164K, and then apply self-training in the rest of the training processes. More details are in \cref{tab:cfg}.

\subsection{Ablation Analysis}
\label{sec:ab}

In this section, we conduct ablation studies to show the effectiveness and respective role of each proposed design.

\subsubsection{Ablation of Framework.}

\revise{Our baseline is the current SOTA one-stage framework \cite{zegclip} of zero-shot semantic segmentation, as shown in \cref{tab:ab}(a).}

\vspace{2mm}\noindent\textbf{\revise{Trusty Token without Learning.}} Firstly, we directly insert the trusty token into the baseline framework. As shown in \cref{tab:ab}(b), the performance on unseen classes falls dramatically due to the poorly-learned token. This gap motivates the design of our Trusty Learner module.

\vspace{2mm}\noindent\textbf{\revise{Trusty Learner.}} A straightforward way to represent the text-image relation is to concatenate text tokens to the image representation. In our work, we propose two improved variants by respectively concatenating the text tokens \rv{$\mathbf{T'}$} before and after inputting them into a multi-head attention block. Two examples are shown in \cref{fig:ab_structure}. \rv{We aim to explore whether it is better for $\mathbf{T'}$ to be directly concatenated with attention or whether it is better to concatenate $\mathbf{T'}$ with $\mathbf{T'}\odot \mathbf{H}$ and then pass through additional attention. Here, we use $\mathbf{T'}\odot \mathbf{H}$ to represent per-element Hadamard product $\mathbf{T'}\odot \mathbf{H}$. In other words, we investigate which approach is more effective: using $\mathbf{T'}$ as a direct concatenation with attention or concatenating $\mathbf{T'}$ with $\mathbf{T'}\odot \mathbf{H}$ and then passing both through the attention.} The experimental results indicate that the latter is superior, specifically the left image in \cref{fig:ab_structure}. 

\vspace{2mm}\noindent\textbf{\revise{Image or Text Representations.}} We also experiment with different representations of image features or image-text relation, including patch embeddings $\mathbf{H}$, \rv{concatenation $[\mathbf{T'}, \mathbf{H}]$, per-element Hadamard product $\mathbf{T'}\odot \mathbf{H}$}, etc. Among the listed experiments, the most impressive performance is achieved by inputting the concatenation (\rv{$\mathbf{\hat{T'}}$}) of the similarity matrix \rv{$\mathbf{T'}\odot \mathbf{H}$} with text tokens \rv{$\mathbf{T'}$} into a self-attention block together, as shown in \cref{tab:ab}(c). This is understandable because the similarity matrix characterizes the text-image relation and the original text tokens are accessible for the module. With this simple design, the trusty token benefits discrimination and boosts the IoU(U) of the baseline by 7.1\%. Furthermore, we weigh the raw segmentation map with the trusty map, which provides an additional 0.3\% improvement on novel classes as shown in \cref{tab:ab}(d).

\subsubsection{Hyper-Parameter $\gamma$.}
In our experimental setup, we adhere to the established convention of using loss weights ${\alpha, \beta}={20, 1}$, a practice consistent with prior research \cite{zegclip, maskformer}. Additionally, we introduced a novel loss term $\mathcal{L}_A$ in \cref{sec:train} as a part of our contribution. To determine the appropriate weight $\gamma$ for this new loss term, we conducted experiments as depicted in \cref{fig:loss_weight}. Our experimental findings revealed that setting $\gamma=10$ resulted in optimal performance. Specifically, on the PASCAL VOC 2012 dataset, this configuration achieved a remarkable mIoU of 93.5\% for seen classes and 85.2\% for unseen classes. Consequently, we have adopted the value $\gamma=10$ as a part of our experimental settings. This careful selection of loss weights underlines our commitment to fine-tuning our approach for the best possible performance, ensuring the robustness and efficacy of our proposed method.

\subsubsection{Initialization}
In our experiments, we initialize the trusty token using full zero initialization. We conducted experiments to investigate the impact of different initialization strategies on the performance of the trusty token. Specifically, we performed three testing iterations on the Pascal VOC dataset and analyzed the results in \cref{tab:init}. Our observations show that initializing the trusty token with full zeros resulted in a performance improvement of 2.5\% compared to initializing it with random values. The full zero initialization strategy seems to provide a more advantageous starting point for the optimization process, leading to better convergence and improved performance in the final results. \revise{These findings contribute to our understanding of the model's sensitivity to initialization and offer valuable insights for achieving improved results in zero-shot semantic segmentation tasks.}

\subsection{Results}
\label{sec:results}
In this section, we compare our proposed TagCLIP with previous approaches.

\subsubsection{Inductive Task.} \revise{First, we evaluate our method for the \textit{inductive} zero-shot segmentation setting, where both names and images of unseen classes are not accessible during training.} Under this setting, our TagCLIP outperforms the mIoU(U) of previous works with significant margins of 7.4\% on PASCAL VOC 2012, and of 1.4\% on COCO-Stuff 164K. Detailed results are in \cref{tab:results}. The better performance on both unseen and seen classes is consistent with our motivation to improve the model's discrimination ability.

\vspace{2mm}\noindent\textbf{Class-wised performance.} \revise{We list the class-wised performance of seen classes in \cref{fig:1_class}(a) and unseen classes in \cref{fig:1_class}(b) on PASCAL VOC 2012.} It shows that TagCLIP beats the current SoTA of every novel class on PASCAL VOC 2012. The margin is especially larger on hard classes, such as the \textit{tvmonitor}, where TagCLIP outperforms the current SoTA ZegCLIP \cite{zegclip} by 30.4\% to 86.7\%. The outstanding performance on unseen categories demonstrates the generalization ability of our TagCLIP. 


\vspace{2mm}\noindent\textbf{Visualization.} We also illustrate the segmentation visualization results in \cref{fig:vis_exp}. \revise{Compared with current SoTA, TagCLIP correctly separates the hard unknown classes, like \textit{cloud} with \textit{sky-other}, and \textit{playingfield} with \textit{dirt}, etc. More visualization results are in the Appendix.}

\vspace{2mm}\noindent\textbf{Parameters and FPS.} In \cref{tab:param}, we present a comparison of the FPS and parameters between our method and ZegCLIP (SoTA). \revise{We conduct evaluations on the same hardware (NVIDIA GeForce RTX 3090) for both the official ZegCLIP implementation and our method to ensure a fair comparison. As a result, our method achieves a significant improvement in FPS (approximately 50\%) and saves 15\% of parameters by simplifying and improving the Segmentor component. We observe that in ZegCLIP \cite{zegclip}, the segmentor block structure follows SegViT\cite{segvit} and includes two attention mechanisms. In contrast, we simplified the structure to include only one attention. This results in parameter savings and a significant increase in FPS. In comparison, the introduction of learnable tokens adds a negligible amount of parameters. By removing redundant attention blocks, we introduce fewer parameters (11.7M) compared to vanilla Segmentor, resulting in enhanced efficiency without compromising performance over ZegCLIP.} This comparison highlights the efficiency and benefits of our method in terms of model complexity.

\re{
\subsubsection{Background Undifferentiated Task}
\label{sec:bg}
As discussed in \cref{sec:results}, we adhered to the experimental setup of previous works \cite{spnet, zegclip}, where novel classes are labeled as an unknown class during training, distinct from a \emph{background} class. To evaluate the generality of our method, we tested it in another setting \cite{zs3} where novel classes are labeled as the background class. This scenario does not allow differentiation between novel classes and the background. Specifically, building on the approach in \cref{sec:results}, we treated the background as an unseen class, which increased the difficulty of segmentation. As shown in \cref{tab:bg}, our method outperformed ZegCLIP by 2.54\%.This demonstrates the robustness of our method across different settings.
}

\subsubsection{Cross-Dataset Task.}
\label{sec:cross_dataset}
To further explore the cross-domain generalization ability of our approach, we conduct extra experiments in \cref{tab:cross_dataset}. We train the model on seen classes of COCO-Stuff 164K and evaluate it on PASCAL Context. Outperforming the current SoTA by 1.4\%, TagCLIP shows better cross-dataset generalization capability. \revise{This also implies that the trusty map remains applicable when all values are 0. This is understandable because when the trusty map classifies as an unseen class if the raw mask originally belongs to an unseen class, it does not compromise the accuracy of unseen class classification. However, suppose the raw mask is erroneously classified as a seen class. In that case, the trusty map will reduce the probability of the seen class and increase the probability of the unseen class, thereby achieving correction.}

\subsubsection{Transductive Task.} 
\label{sec:trans}
Besides, we evaluate our method for another setting called \textit{transductive} zero-shot learning \cite{gagnet, denseclip}. It allows access to unseen category names during training, but their ground truth masks remained unavailable. TagCLIP is not designed for the \textit{transductive} setting because self-supervision training on accessible unseen names \cite{zegclip, denseclip, spnet} serves the same role. Despite this, TagCLIP still excels the mIoU(U) of the current SoTA \cite{zegclip} by 2.8\% on PASCAL VOC 2012 and reaches comparative performance on COCO-Stuff 164K. Results are in \cref{tab:transductive}. It demonstrates TagCLIP's superior generalization ability to different tasks.

\subsubsection{Fully Supervised Task.} 
\label{sec:full}
We also provide the results trained with full labels in \cref{tab:supervision}. \revise{It demonstrates that TagCLIP improves the upper bound of zero-shot segmentation results by 1.9\% on PASCAL VOC 2012 and by 0.4\% on COCO-Stuff 164K. Similar to \cref{sec:cross_dataset}, when the trusty map classifies as a seen class, if the raw mask was originally classified as a seen class, there will be no compromise to the accuracy of the seen class classification. However, suppose the raw mask was originally misclassified as an unseen class. In that case, the trusty map will decrease the probability of the unseen class and increase the probability of the seen class, achieving the correction objective.}

%% file: sections/08_conclusion.tex
\revise{In this work, we have proposed a novel framework, TagCLIP (Trusty-aware guided CLIP), to address the long-standing issue of recognizing unseen classes in zero-shot semantic segmentation.} By disentangling the baseline into two parallel processes, one for semantic matching and the other for predicting reliability, we have designed a trusty token that captures the prediction tendency of known and novel categories. With almost no extra overhead from the disentangled objectives, TagCLIP effectively upgrades the pixel-level generalization capacity of existing models. Our experimental results demonstrate the effectiveness of TagCLIP, as it outperforms state-of-the-art approaches by 7.4\% on PASCAL VOC 2012 and 1.7\% on COCO-Stuff 164K.

%% file: sections/09_bio.tex
\begin{IEEEbiography}
 [{\includegraphics[height=1.10in,clip,keepaspectratio]{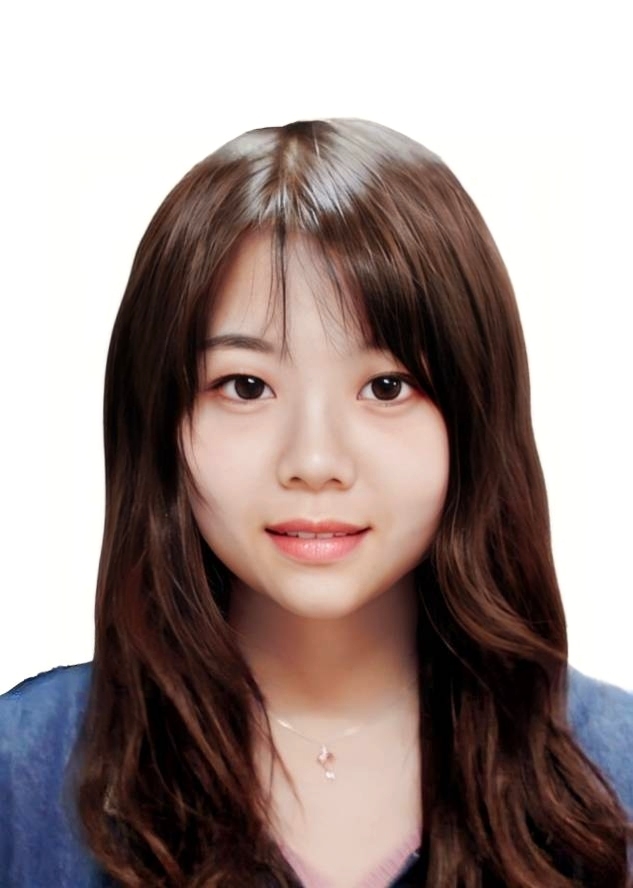}}]
 {Jingyao Li} received the B.Eng. degree from Xi'an Jiaotong University. She is currently a Ph.D. student at Department of Computer Science and Engineering of the Chinese University of Hong Kong (CUHK), under the supervision of Prof. Jiaya Jia. Her research interests include self-supervised learning, knowledge distillation and out-of-distribution detection.
\end{IEEEbiography}

\begin{IEEEbiography}
 [{\includegraphics[height=1.25in,clip,keepaspectratio]{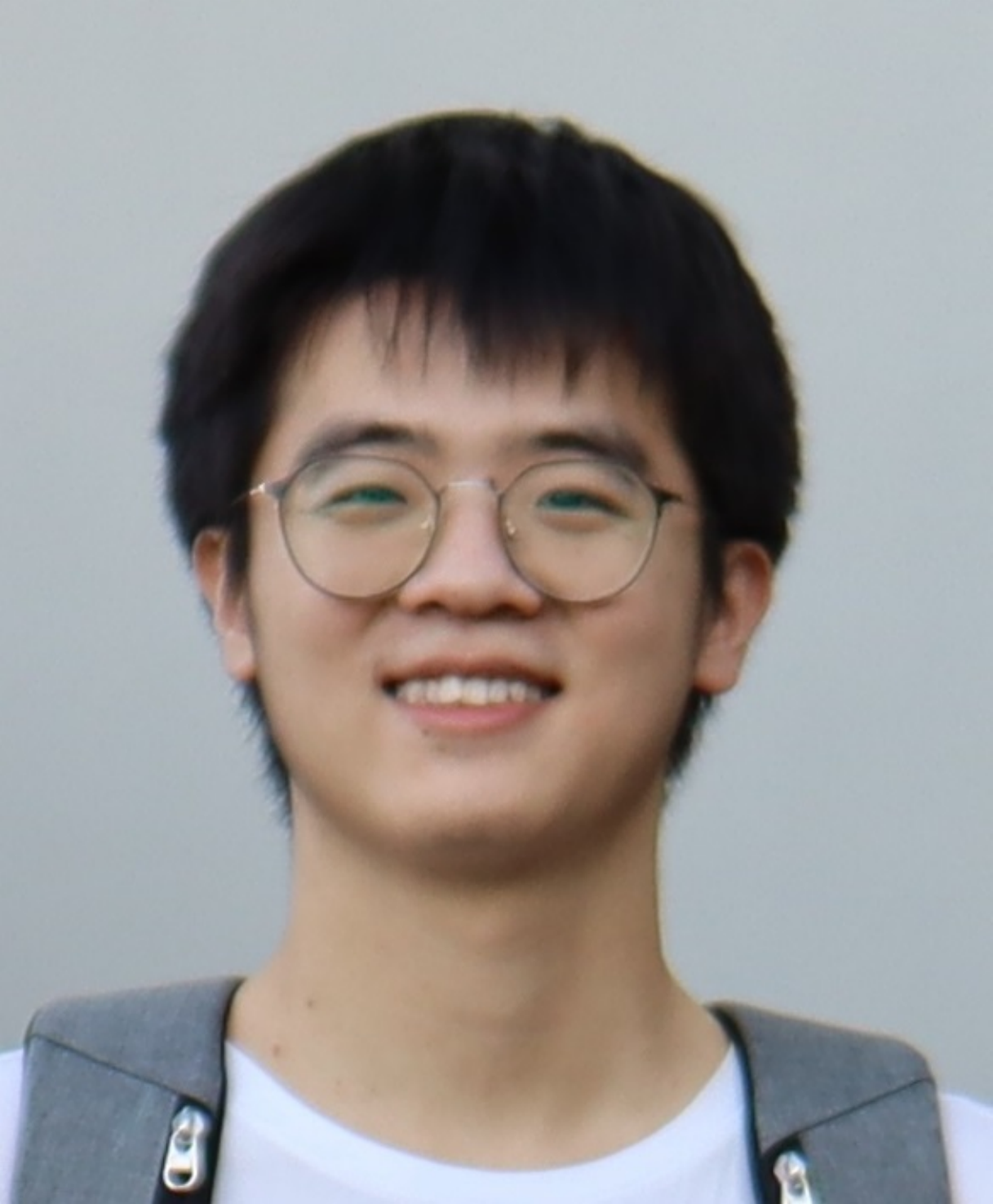}}] {Pengguang Chen} received the B.Eng. degree in Computer Science from Nanjing University and the Ph.D. degree from the Chinese University of Hong Kong (CUHK), under the supervision of Prof. Jiaya Jia. He is currently a researcher in SmartMore. He serves as a reviewer for CVPR, ICCV, ECCV, TPAMI. His research interests include neural architecture search, self-supervised learning, knowledge distillation and semantic segmentation.
\end{IEEEbiography}

\begin{IEEEbiography}[{\includegraphics[width=1in,height=1.25in,clip,]{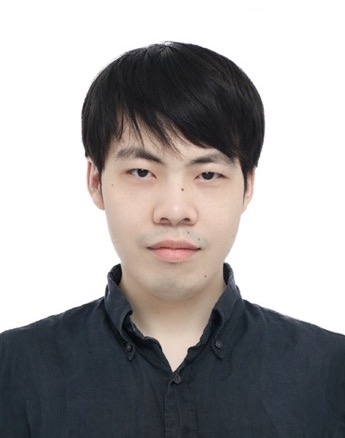}}] {Shengju Qian} received his Ph.D. degree from The Chinese University of Hong Kong, supervised by Prof. Jiaya Jia. He received his Bachelor degree from the University of Electronic Science and Technology of China. He serves as a reviewer for TPAMI, CVPR, ICCV, NeurIPS, ECCV, etc. His research interests lie primarily in computer vision and deep learning.
\end{IEEEbiography}

\begin{IEEEbiography}
[{\includegraphics[width=1in,height=1.25in,clip,keepaspectratio]{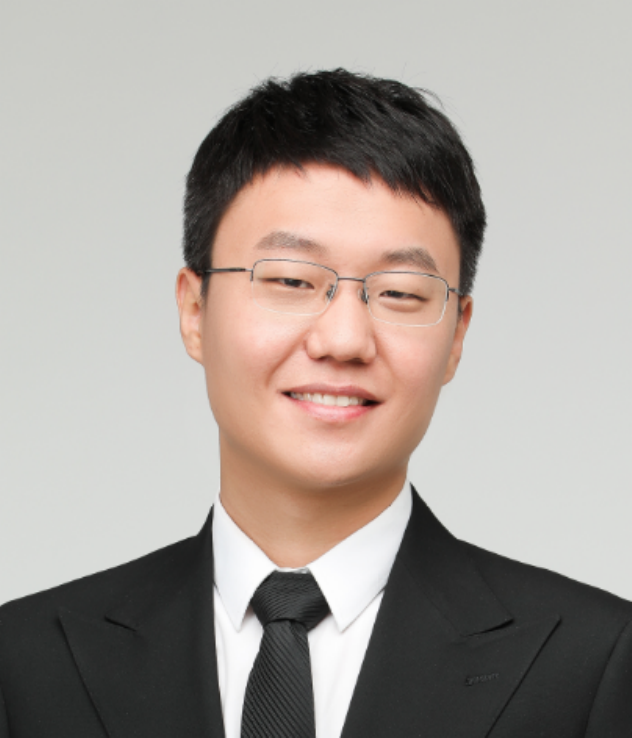}}]{Shu Liu} now serves as Co-Founder and Technical Head in SmartMore. He received the BS degree from Huazhong University of Science and Technology and the PhD degree from the Chinese University of Hong Kong. He was the winner of 2017 COCO Instance Segmentation Competition and received the Outstanding Reviewer of ICCV in 2019. He continuously served as a reviewer for TPAMI, CVPR, ICCV, NIPS, ICLR and etc. His research interests lie in deep learning and computer vision. 
  He is a member of IEEE.
 \end{IEEEbiography} 

\begin{IEEEbiography}
[{\includegraphics[width=1in,height=1.25in,clip,keepaspectratio]{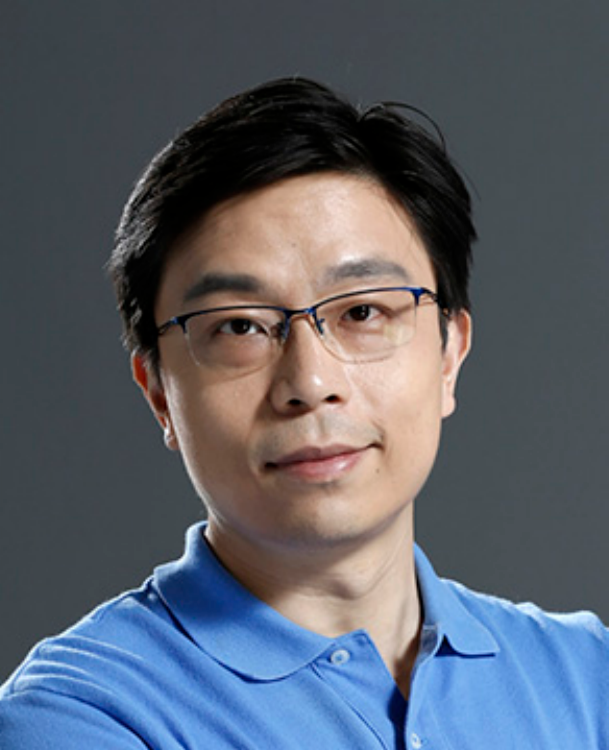}}]{Jiaya Jia} received the Ph.D.~degree in Computer Science from Hong Kong University of Science and Technology in 2004 and is currently a full professor in Department of Computer Science and Engineering at the Chinese University of Hong Kong (CUHK). He assumes the position of Associate Editor-in-Chief of IEEE Transactions on Pattern Analysis and Machine Intelligence (TPAMI) and is in the editorial board of International Journal of Computer Vision (IJCV). He continuously served as area chairs for ICCV, CVPR, AAAI, ECCV, and several other conferences for the organization. He was on program committees of major conferences in graphics and computational imaging, including ICCP, SIGGRAPH, and SIGGRAPH Asia. He is a Fellow of the IEEE. 
\end{IEEEbiography}

%% file: main_paper.bbl
\begin{thebibliography}{10}
\providecommand{\url}[1]{#1}
\csname url@samestyle\endcsname
\providecommand{\newblock}{\relax}
\providecommand{\bibinfo}[2]{#2}
\providecommand{\BIBentrySTDinterwordspacing}{\spaceskip=0pt\relax}
\providecommand{\BIBentryALTinterwordstretchfactor}{4}
\providecommand{\BIBentryALTinterwordspacing}{\spaceskip=\fontdimen2\font plus
\BIBentryALTinterwordstretchfactor\fontdimen3\font minus \fontdimen4\font\relax}
\providecommand{\BIBforeignlanguage}[2]{{%
\expandafter\ifx\csname l@#1\endcsname\relax
\typeout{** WARNING: IEEEtran.bst: No hyphenation pattern has been}%
\typeout{** loaded for the language `#1'. Using the pattern for}%
\typeout{** the default language instead.}%
\else
\language=\csname l@#1\endcsname
\fi
#2}}
\providecommand{\BIBdecl}{\relax}
\BIBdecl

\bibitem{mood}
J.~Li, P.~Chen, S.~Yu, Z.~He, S.~Liu, and J.~Jia, ``Rethinking out-of-distribution (ood) detection: Masked image modeling is all you need,'' \emph{arXiv preprint arXiv:2302.02615}, 2023.

\bibitem{motcoder}
J.~Li, P.~Chen, and J.~Jia, ``Motcoder: Elevating large language models with modular of thought for challenging programming tasks,'' 2024.

\bibitem{moodv2}
J.~Li, P.~Chen, S.~Yu, S.~Liu, and J.~Jia, ``Moodv2: Masked image modeling for out-of-distribution detection,'' 2024.

\bibitem{bal}
------, ``Bal: Balancing diversity and novelty for active learning,'' \emph{IEEE Transactions on Pattern Analysis and Machine Intelligence}, pp. 1--12, 2023.

\bibitem{ovcls}
H.~Pham, Z.~Dai, G.~Ghiasi, K.~Kawaguchi, H.~Liu, A.~W. Yu, J.~Yu, Y.-T. Chen, M.-T. Luong, Y.~Wu \emph{et~al.}, ``Combined scaling for open-vocabulary image classification,'' \emph{arXiv e-prints}, pp. arXiv--2111, 2021.

\bibitem{ovseg}
F.~Liang, B.~Wu, X.~Dai, K.~Li, Y.~Zhao, H.~Zhang, P.~Zhang, P.~Vajda, and D.~Marculescu, ``Open-vocabulary semantic segmentation with mask-adapted clip,'' \emph{arXiv preprint arXiv:2210.04150}, 2022.

\bibitem{baseline}
M.~Xu, Z.~Zhang, F.~Wei, Y.~Lin, Y.~Cao, H.~Hu, and X.~Bai, ``A simple baseline for zero-shot semantic segmentation with pre-trained vision-language model,'' \emph{arXiv preprint arXiv:2112.14757}, 2021.

\bibitem{zegclip}
Z.~Zhou, B.~Zhang, Y.~Lei, L.~Liu, and Y.~Liu, ``Zegclip: Towards adapting clip for zero-shot semantic segmentation,'' \emph{arXiv preprint arXiv:2212.03588}, 2022.

\bibitem{ovdet}
Y.~Du, F.~Wei, Z.~Zhang, M.~Shi, Y.~Gao, and G.~Li, ``Learning to prompt for open-vocabulary object detection with vision-language model,'' in \emph{Proceedings of the IEEE/CVF Conference on Computer Vision and Pattern Recognition}, 2022, pp. 14\,084--14\,093.

\bibitem{ovdet_kd}
X.~Gu, T.-Y. Lin, W.~Kuo, and Y.~Cui, ``Open-vocabulary object detection via vision and language knowledge distillation,'' \emph{arXiv preprint arXiv:2104.13921}, 2021.

\bibitem{fcn}
J.~Long, E.~Shelhamer, and T.~Darrell, ``Fully convolutional networks for semantic segmentation,'' in \emph{Proceedings of the IEEE conference on computer vision and pattern recognition}, 2015, pp. 3431--3440.

\bibitem{unet}
O.~Ronneberger, P.~Fischer, and T.~Brox, ``U-net: Convolutional networks for biomedical image segmentation,'' in \emph{Medical Image Computing and Computer-Assisted Intervention--MICCAI 2015: 18th International Conference, Munich, Germany, October 5-9, 2015, Proceedings, Part III 18}.\hskip 1em plus 0.5em minus 0.4em\relax Springer, 2015, pp. 234--241.

\bibitem{deeplab}
L.-C. Chen, G.~Papandreou, I.~Kokkinos, K.~Murphy, and A.~L. Yuille, ``Semantic image segmentation with deep convolutional nets and fully connected crfs,'' \emph{arXiv preprint arXiv:1412.7062}, 2014.

\bibitem{deeplabv2}
L.-C. Chen, G.~Papandreou, F.~Schroff, and H.~Adam, ``Rethinking atrous convolution for semantic image segmentation,'' \emph{arXiv preprint arXiv:1706.05587}, 2017.

\bibitem{clip}
A.~Radford, J.~W. Kim, C.~Hallacy, A.~Ramesh, G.~Goh, S.~Agarwal, G.~Sastry, A.~Askell, P.~Mishkin, J.~Clark \emph{et~al.}, ``Learning transferable visual models from natural language supervision,'' in \emph{International conference on machine learning}.\hskip 1em plus 0.5em minus 0.4em\relax PMLR, 2021, pp. 8748--8763.

\bibitem{feng2020language}
F.~Feng, Y.~Yang, D.~Cer, N.~Arivazhagan, and W.~Wang, ``Language-agnostic bert sentence embedding,'' \emph{arXiv preprint arXiv:2007.01852}, 2020.

\bibitem{takahashi2022unsupervised}
K.~Takahashi and D.~Bollegala, ``Unsupervised attention-based sentence-level meta-embeddings from contextualised language models,'' \emph{arXiv preprint arXiv:2204.07746}, 2022.

\bibitem{bert}
J.~Devlin, M.-W. Chang, K.~Lee, and K.~Toutanova, ``Bert: Pre-training of deep bidirectional transformers for language understanding,'' 2019.

\bibitem{align}
C.~Jia, Y.~Yang, Y.~Xia, Y.-T. Chen, Z.~Parekh, H.~Pham, Q.~Le, Y.-H. Sung, Z.~Li, and T.~Duerig, ``Scaling up visual and vision-language representation learning with noisy text supervision,'' in \emph{International Conference on Machine Learning}.\hskip 1em plus 0.5em minus 0.4em\relax PMLR, 2021, pp. 4904--4916.

\bibitem{visualbert}
L.~H. Li, M.~Yatskar, D.~Yin, C.-J. Hsieh, and K.-W. Chang, ``Visualbert: A simple and performant baseline for vision and language,'' \emph{arXiv preprint arXiv:1908.03557}, 2019.

\bibitem{vlbert}
W.~Su, X.~Zhu, Y.~Cao, B.~Li, L.~Lu, F.~Wei, and J.~Dai, ``Vl-bert: Pre-training of generic visual-linguistic representations,'' \emph{arXiv preprint arXiv:1908.08530}, 2019.

\bibitem{liu2021image}
Z.~Liu, C.~Rodriguez-Opazo, D.~Teney, and S.~Gould, ``Image retrieval on real-life images with pre-trained vision-and-language models,'' in \emph{Proceedings of the IEEE/CVF International Conference on Computer Vision}, 2021, pp. 2125--2134.

\bibitem{jiang2022finetuning}
J.~Jiang, Z.~Liu, and N.~Zheng, ``Finetuning pretrained vision-language models with correlation information bottleneck for robust visual question answering,'' \emph{arXiv preprint arXiv:2209.06954}, 2022.

\bibitem{cris}
Z.~Wang, Y.~Lu, Q.~Li, X.~Tao, Y.~Guo, M.~Gong, and T.~Liu, ``Cris: Clip-driven referring image segmentation,'' in \emph{Proceedings of the IEEE/CVF conference on computer vision and pattern recognition}, 2022, pp. 11\,686--11\,695.

\bibitem{denseclip}
C.~Zhou, C.~C. Loy, and B.~Dai, ``Denseclip: Extract free dense labels from clip,'' \emph{arXiv preprint arXiv:2112.01071}, 2021.

\bibitem{zhang2023simple}
H.~Zhang, F.~Li, X.~Zou, S.~Liu, C.~Li, J.~Yang, and L.~Zhang, ``A simple framework for open-vocabulary segmentation and detection,'' in \emph{Proceedings of the IEEE/CVF International Conference on Computer Vision}, 2023, pp. 1020--1031.

\bibitem{ghiasi2022scaling}
G.~Ghiasi, X.~Gu, Y.~Cui, and T.-Y. Lin, ``Scaling open-vocabulary image segmentation with image-level labels,'' in \emph{European Conference on Computer Vision}.\hskip 1em plus 0.5em minus 0.4em\relax Springer, 2022, pp. 540--557.

\bibitem{qin2023freeseg}
J.~Qin, J.~Wu, P.~Yan, M.~Li, R.~Yuxi, X.~Xiao, Y.~Wang, R.~Wang, S.~Wen, X.~Pan \emph{et~al.}, ``Freeseg: Unified, universal and open-vocabulary image segmentation,'' in \emph{Proceedings of the IEEE/CVF Conference on Computer Vision and Pattern Recognition}, 2023, pp. 19\,446--19\,455.

\bibitem{segmenter}
R.~Strudel, R.~Garcia, I.~Laptev, and C.~Schmid, ``Segmenter: Transformer for semantic segmentation,'' in \emph{Proceedings of the IEEE/CVF international conference on computer vision}, 2021, pp. 7262--7272.

\bibitem{segformer}
E.~Xie, W.~Wang, Z.~Yu, A.~Anandkumar, J.~M. Alvarez, and P.~Luo, ``Segformer: Simple and efficient design for semantic segmentation with transformers,'' \emph{Advances in Neural Information Processing Systems}, vol.~34, pp. 12\,077--12\,090, 2021.

\bibitem{encnet}
H.~Zhang, K.~Dana, J.~Shi, Z.~Zhang, X.~Wang, A.~Tyagi, and A.~Agrawal, ``Context encoding for semantic segmentation,'' in \emph{Proceedings of the IEEE conference on Computer Vision and Pattern Recognition}, 2018, pp. 7151--7160.

\bibitem{sert}
S.~Zheng, J.~Lu, H.~Zhao, X.~Zhu, Z.~Luo, Y.~Wang, Y.~Fu, J.~Feng, T.~Xiang, P.~H. Torr \emph{et~al.}, ``Rethinking semantic segmentation from a sequence-to-sequence perspective with transformers,'' in \emph{Proceedings of the IEEE/CVF conference on computer vision and pattern recognition}, 2021, pp. 6881--6890.

\bibitem{mask2former}
B.~Cheng, I.~Misra, A.~G. Schwing, A.~Kirillov, and R.~Girdhar, ``Masked-attention mask transformer for universal image segmentation,'' in \emph{Proceedings of the IEEE/CVF Conference on Computer Vision and Pattern Recognition}, 2022, pp. 1290--1299.

\bibitem{maskformer}
B.~Cheng, A.~Schwing, and A.~Kirillov, ``Per-pixel classification is not all you need for semantic segmentation,'' \emph{Advances in Neural Information Processing Systems}, vol.~34, pp. 17\,864--17\,875, 2021.

\bibitem{segvit}
B.~Zhang, Z.~Tian, Q.~Tang, X.~Chu, X.~Wei, C.~Shen \emph{et~al.}, ``Segvit: Semantic segmentation with plain vision transformers,'' \emph{Advances in Neural Information Processing Systems}, vol.~35, pp. 4971--4982, 2022.

\bibitem{spnet}
Y.~Xian, S.~Choudhury, Y.~He, B.~Schiele, and Z.~Akata, ``Semantic projection network for zero-and few-label semantic segmentation,'' in \emph{Proceedings of the IEEE/CVF Conference on Computer Vision and Pattern Recognition}, 2019, pp. 8256--8265.

\bibitem{zs3}
M.~Bucher, T.-H. Vu, M.~Cord, and P.~P{\'e}rez, ``Zero-shot semantic segmentation,'' \emph{Advances in Neural Information Processing Systems}, vol.~32, 2019.

\bibitem{gagnet}
Z.~Gu, S.~Zhou, L.~Niu, Z.~Zhao, and L.~Zhang, ``Context-aware feature generation for zero-shot semantic segmentation,'' in \emph{Proceedings of the 28th ACM International Conference on Multimedia}, 2020, pp. 1921--1929.

\bibitem{sign}
J.~Cheng, S.~Nandi, P.~Natarajan, and W.~Abd-Almageed, ``Sign: Spatial-information incorporated generative network for generalized zero-shot semantic segmentation,'' in \emph{Proceedings of the IEEE/CVF International Conference on Computer Vision}, 2021, pp. 9556--9566.

\bibitem{joint}
D.~Baek, Y.~Oh, and B.~Ham, ``Exploiting a joint embedding space for generalized zero-shot semantic segmentation,'' in \emph{Proceedings of the IEEE/CVF international conference on computer vision}, 2021, pp. 9536--9545.

\bibitem{strict}
G.~Pastore, F.~Cermelli, Y.~Xian, M.~Mancini, Z.~Akata, and B.~Caputo, ``A closer look at self-training for zero-label semantic segmentation,'' in \emph{Proceedings of the IEEE/CVF Conference on Computer Vision and Pattern Recognition}, 2021, pp. 2693--2702.

\bibitem{dice}
F.~Milletari, N.~Navab, and S.-A. Ahmadi, ``V-net: Fully convolutional neural networks for volumetric medical image segmentation,'' in \emph{2016 fourth international conference on 3D vision (3DV)}.\hskip 1em plus 0.5em minus 0.4em\relax Ieee, 2016, pp. 565--571.

\bibitem{focal}
T.-Y. Lin, P.~Goyal, R.~Girshick, K.~He, and P.~Doll{\'a}r, ``Focal loss for dense object detection,'' in \emph{Proceedings of the IEEE international conference on computer vision}, 2017, pp. 2980--2988.

\bibitem{word2vec}
T.~Mikolov, I.~Sutskever, K.~Chen, G.~Corrado, and J.~Dean, ``Distributed representations of words and phrases and their compositionality,'' 2013.

\bibitem{fasttext}
A.~Joulin, E.~Grave, P.~Bojanowski, M.~Douze, H.~J{\'e}gou, and T.~Mikolov, ``Fasttext.zip: Compressing text classification models,'' \emph{arXiv preprint arXiv:1612.03651}, 2016.

\bibitem{zegformer}
J.~Ding, N.~Xue, G.-S. Xia, and D.~Dai, ``Decoupling zero-shot semantic segmentation,'' in \emph{Proceedings of the IEEE/CVF Conference on Computer Vision and Pattern Recognition}, 2022, pp. 11\,583--11\,592.

\end{thebibliography}
